\documentclass{article}

\usepackage{microtype}
\usepackage{graphicx}
\usepackage{subfigure}
\usepackage{booktabs} 
\usepackage{amsmath}
\usepackage{amssymb}
\usepackage{bm}
\usepackage{enumitem} 

\usepackage{siunitx}  
\usepackage{hyperref}

\usepackage[accepted]{icml2018}

\icmltitlerunning{Learning by Playing}

\usepackage[mathscr]{euscript}

\newcommand{\RLD}{Reinforcement Learning (RL) }
\newcommand{\MDPD}{Markov Decision Process (MDP) }

\newcommand{\bE}{\mathbb{E}}
\newcommand{\bx}{\mathbf{x}}
\newcommand{\bs}{\mathbf{s}}
\newcommand{\ba}{\mathbf{a}}
\newcommand{\btheta}{{\bm{\theta}}}
\newcommand{\cM}{{\mathcal{M}}}
\newcommand{\sA}{\mathscr{A}}
\newcommand{\sT}{\mathscr{T}}
\newcommand{\cA}{{\mathcal{A}}}
\newcommand{\cT}{{\mathcal{T}}}
\newcommand{\cB}{{\mathcal{B}}}
\newcommand{\cL}{{\mathcal{L}}}
\newcommand{\cS}{{\mathcal{S}}}
\newcommand{\pluseq}{\mathrel{+}=}

\begin{document}

\twocolumn[
\icmltitle{Learning by Playing -- Solving Sparse Reward Tasks from Scratch}




\icmlsetsymbol{equal}{*}

\begin{icmlauthorlist}
\icmlauthor{Martin Riedmiller}{equal,goo}
\icmlauthor{Roland Hafner}{equal,goo}
\icmlauthor{Thomas Lampe}{goo}
\icmlauthor{Michael Neunert}{goo}
\icmlauthor{Jonas Degrave}{goo}
\icmlauthor{Tom Van de Wiele}{goo}
\icmlauthor{Volodymyr Mnih}{goo}
\icmlauthor{Nicolas Heess}{goo}
\icmlauthor{Tobias Springenberg}{goo}
\end{icmlauthorlist}

\icmlaffiliation{goo}{Google DeepMind, London, GB}

\icmlcorrespondingauthor{Martin Riedmiller}{riedmiller@google.com}
\icmlcorrespondingauthor{Roland Hafner}{rhafner@google.com}

\icmlkeywords{Reinforcement Learning, Manipulation, Robot, Machine Learning, ICML}

\vskip 0.3in
]

\printAffiliationsAndNotice{\icmlEqualContribution} 

\begin{abstract}
We propose Scheduled Auxiliary Control (SAC-X), a new learning paradigm in the context of Reinforcement Learning (RL). SAC-X enables learning of complex behaviors -- from scratch -- in the presence of multiple sparse reward signals.
To this end, the agent is equipped with a set of general auxiliary tasks, that it attempts to learn simultaneously via off-policy RL.
The key idea behind our method is that active (learned) scheduling and execution of auxiliary policies allows the agent to efficiently explore its environment -- enabling it to excel at sparse reward RL.
Our experiments in several challenging robotic manipulation settings demonstrate the power of our approach. A video of the rich set of learned behaviours can be found at \href{https://youtu.be/mPKyvocNe_M}{https://youtu.be/mPKyvocNe\_M}.
\end{abstract}

\section{Introduction}
\label{sec:intro}

Consider the following scenario: a learning agent has to control a robot arm to open a box and place a block inside.
While defining the reward for this task is simple and straightforward (e.g. using a simple mechanism inside the box such as a force sensor to detect a placed block), the underlying learning problem is hard.
The agent has to discover a long sequence of ``correct'' actions in order to find a configuration of the environment that yields the sparse reward -- the block contained inside the box. Discovering this sparse reward signal is a hard exploration problem for which success via random exploration is highly unlikely. 

Over the last decades, a multitude of methods have been developed to help with the above mentioned exploration problem.
These include for example: shaping rewards~\citep{Ng19,Randlov98,Gu17}, curriculum learning~\citep{HeessParkour17,Ghosh18,Oudeyer17}, transfer of learned policies from simulation to reality (see \citet{Duan17,Sadeghi17,Tobin2017,Rusu17} for recent examples), learning from demonstrations~\citep{Ross11,Vecerik17,Kober2011,Sermanet17,Nair17}, learning with model guidance, see e.g.  \citet{Montgomery16}, or inverse RL~\citep{Ng00,Ziebart08}. 
All of these approaches rely on the availability of prior knowledge that is specific to a task. Moreover, they often bias the control policy in a certain -- potentially suboptimal -- direction. For example, using a shaped reward designed by the experimenter, inevitably, biases the solutions that the agent can find. In contrast to this, when a sparse task formulation is used, the agent can discover novel and potentially preferable solutions. 
We would thus, arguably, prefer to develop methods that support the agent during learning but preserve the ability of the agent to learn from sparse rewards. Ideally, our new methods should reduce the specific prior task knowledge that is required to cope with sparse rewards.

In this paper, we introduce a new method dubbed Scheduled Auxiliary Control (SAC-X), as a first step towards such an approach. It is based on four main principles: 
\begin{enumerate}[noitemsep,nolistsep]
    \item Every state-action pair is paired with a vector of rewards, consisting of (typically sparse) externally provided rewards and (typically sparse) internal auxiliary rewards.
    \item Each reward entry has an assigned policy, called intention in the following, which is trained to maximize its corresponding cumulative reward.
    \item There is a high-level scheduler which selects and executes the individual intentions with the goal of improving performance of the agent on the external tasks.
    \item Learning is performed off-policy (and asynchronously from policy execution) and the experience between intentions is shared -- to use information effectively.
\end{enumerate}

Although the approach proposed in this paper is generally applicable to a wider range of problems, we discuss our method in the light of a typical robotics manipulation application with sparse rewards: stacking various objects and cleaning a table.

Auxiliary rewards in these tasks are defined based on the mastery of the agent to control its own sensory observations (e.g. images, proprioception, haptic sensors). They are designed to be easily implementable in a real robot setup. In particular, we define auxiliary rewards on a raw sensory level -- e.g. whether a touch is detected or not. Or, alternatively, define them on a higher level that requires a small amount of pre-computation of entities, e.g. whether any object moved or whether two objects are close to each other in the image plane. Based on these basic auxiliary tasks, the agent must effectively explore its environment until more interesting, external rewards are observed; an approach which is inspired by the playful phase of childhood in humans.

We demonstrate the capabilities of SAC-X in simulation on challenging robot manipulation tasks such as stacking and tidying a table-top using a robot arm. 
All tasks are defined via sparse, easy to define, rewards and solved using the same set of auxiliary reward functions.
In addition, we demonstrate that our method is sample efficient, allowing us to learn from scratch on a real robot.

\section{Related Work}
\label{sec:related}

The idea of using auxiliary tasks in the context of reinforcement learning has been explored several times in literature. 
Among the first papers to make use of this idea, is the work by \cite{Sutton2011horde} where general value functions are learned for a large collection of pseudo-rewards corresponding to different goals extracted from the sensorimotor stream.
General value functions have recently been extended to Deep RL in work on Universal Value Function Aproximators (UVFA) \citep{Schaul15}. These are in turn are inherently connected to learning to predict the future via ``successor'' representations \citep{Dayan93,Kulkarni2016,Barreto2017} or forecasts \citep{Schaul2013forecasts,Lample17,Dosovitskiy2017} and are trained to be predictive of features extracted from future states.
In contrast to the setting explored in this paper, all of the aforementioned approaches do not utilize the learned sub-policies to drive exploration for an external ``common'' goal. They also typically assume independence between different policies and value functions.
In a similar vein to the UVFA approach, recent work on Hindsight Experience Replay (HER) \cite{andrychowicz2017hindsight} proposed to generate many tasks for a reinforcement learning agent by randomly sampling goals along previously experienced trajectories. Our approach can be understood as an extension of HER to semantically grounded, and scheduled, goals.

A related strand of research has considered learning a shared representation for multiple RL tasks.

Closest to the ideas presented in this paper and serving as the main inspiration for our approach, is the recent work on Deep Reinforcement Learning with the UNREAL agent \citep{Jaderberg2017Unreal} and Actor Critic agents for navigation \citep{Mirowski16} (discrete action control) as well as the Intentional Unintentional Agent \citep{cabi2017} (considering continuous actions). While these approaches mainly consider using auxiliary tasks to provide additional learning signals -- and additional exploration by following random sensory goals -- we here make active use of the auxiliary tasks by switching between them throughout individual episodes (to achieve exploration for the main task). 

Our work is also connected to the broader literature on multi-task (reinforcement) learning (see e.g. \citet{Caruana97} for a general overview and \citet{Lazaric2008,Mehta08} for RL applications) and work on reinforcement learning via options \citep{Dietterich98, Bacon17, DanielNP2012}. In contrast to these approaches we here learn skills that are semantically grounded via auxiliary rewards, instead of automatically discovering a decomposed solution to a single task.

The approach we take for scheduling the learning and execution of different auxiliary tasks can be understood from the perspective of ``teaching'' a set of increasingly more complicated problems -- see e.g. the literature on curriculum learning \citep{Bengio2009} -- where we consider a fixed number of problems and learn a teaching policy online. Research on this topic has a long history, both in the machine learning and psychology literature. Recent examples from the field of RL include the PowerPlay algorithm \citep{Schmidhuber2013PowerPlayTA}, that invents and teaches new problems on the fly, as well as research on learning complex tasks via curriculum learning for RL \citep{HeessParkour17} and hierarchical learning of real robot tasks \citep{Oudeyer17}. 
(Hierarchical) Reinforcement Learning with the help of so called ``intrinsic motivation'' rewards \citep{Chentanez04,Singh09} has, furthermore, been studied for controlling real robots by \citet{Ngo2012} and combined with Deep RL techniques by \citet{KulkarniNST16,DilokthanakulKP17}. In contrast to our work these approaches typically consider internal measures such as learning progress to define rewards, rather than auxiliary tasks that are grounded in physical reality.

\section{Preliminaries}
We consider the problem of \RLD in a \MDPD. We make use of the following basic definitions: Let $\bs \in \mathbb{R}^S$ be the state of the agent in the MDP $\cM$ -- we use the term state and observation of the state (e.g. proprioceptive features, object positions or images) interchangeably to simplify notation. Denote with $\ba \in \mathbb{R}^A$ the action vector and $p(\bs_{t+1} | \bs_t, \ba_t)$ the probability density of transitioning to state $\bs_{t+1}$ when executing action $\ba_t$ in $\bs_t$. 
All actions are assumed to be sampled from a policy distribution $\pi_\btheta(\ba | \bs)$, with parameters $\btheta$. After executing an action -- and transitioning in the environment -- the agent receives a scalar reward $r_\cM(\bs_t, \ba_t)$.

With these definitions in place, we can define the goal of Reinforcement Learning as maximizing the sum of discounted rewards $\bE_{\pi} \lbrack R(\tau_{0:\infty}) \rbrack = \bE_{\pi} \lbrack \sum_{t=0}^\infty \gamma^t r(\bs_t, \ba_t) \mid a_t \sim \pi(\cdot | \bs_t),\, \bs_{t+1} \sim p(\cdot | \bs_t, \ba_t),\, \bs_0 \sim p(\bs)  \rbrack$, where $p(\bs)$ denotes the initial state distribution or, if assuming random restarts, the state visitation distribution, and we use the short notation $\tau_{t:\infty} = \lbrace (\bs_t, \ba_t), \dots \rbrace$ to refer to the trajectory starting in state $t$. For brevity of notation, we will, in the following, omit the dependence of the expectation on samples from the transition and initial state distribution where unambiguous.

\section{Scheduled Auxiliary Control}
We will now introduce our method for RL in sparse reward problems. For the purpose of this paper, we define a sparse reward problem as finding the optimal policy $\pi^*$ in an MDP $\cM$ with a reward function that is characterized by an '$\epsilon$-region' in state space. That is we have 
\begin{equation}
r_\cM(\bs, \ba) = \begin{cases}
 \delta_{\bs_g}(\bs) &\text{if } \ d(\bs, \bs_g) \leq \epsilon \\
 0 &\text{else},
 \end{cases}
 \label{eq:sparse_rew}
\end{equation}
where $\bs_g$ denotes a goal state, $d(\bs, \bs_g)$ denotes the distance between the goal state and the current state $s$ -- defined on a subset of the variables comprising $s$ and measured according to some metric, i.e. we could have $d(\bs, \bs_g) = \| \bs - \bs_g \|_2$. Further, $\delta_{\bs_g}(\bs)$ defines the reward surface within the epsilon region; in this paper we will choose the most extreme case where $\epsilon$ is small and we set $\delta_{\bs_g}(\bs) = 1$ (constant).\footnote{Instead, we could also define a small reward gradient within the $\epsilon$-region to enforce precise control by setting, for example, $\delta_{\bs_g}(\bs) = \exp(-d(\bs, \bs_g))$.}

\subsection{A Hierarchical RL Approach for Learning from Sparse Rewards}
To enable learning in the setting described above we derive an algorithm that augments the sparse learning problem with a set of low-level auxiliary tasks.

Formally, let $\sA = \lbrace\cA_1, \dots, \cA_K\rbrace$ denote the set of auxiliary MDPs. In our construction, these MDPs share the state, observation and action space as well as the transition dynamics with the main task $\cM$,\footnote{We note that in the experiments we later also allow for multiple external (main) tasks, but omit this detail here for clarity of the presentation.} but have separate auxiliary reward functions $r_{\cA_1}(\bs, \ba), \dots, r_{\cA_K}(\bs, \ba)$. We assume full control over the auxiliary rewards; i.e. we assume knowledge of how to compute auxiliary rewards and assume we can evaluate them at any state action pair. Although this assumption might appear restrictive at first glance, we will -- as mentioned before -- make use of simple auxiliary rewards that can be obtained from the activation of the agents sensors.

Given the set of reward functions we can define intention policies and their return as $\pi_\btheta(\ba | \bs, \cT)$ and 
\begin{equation}
\bE_{\pi_\btheta(\ba | \bs, \cT)} \Big\lbrack R_\cT(\tau_{t:\infty}) \Big\rbrack = \bE_{\pi_\btheta(\ba | \bs, \cT)} \Big\lbrack \sum_{t=0}^\infty \gamma^t r_\cT(\bs_t, \ba_t) \Big\rbrack,
\end{equation}
where $\cT \in \sT = \sA \cup \lbrace \cM \rbrace$, respectively.

To derive a learning objective based on these definitions it is useful to first remind ourselves what the aim of such a procedure should be: Our goal for learning is to both, i) train all auxiliary intentions policies and the main task policy to achieve their respective goals, and ii) utilize all intentions for fast exploration in the main sparse-reward MDP $\cM$. We accomplish this by defining a hierarchical objective for policy training that decomposes into two parts. 

\paragraph{Learning the intentions}
The first part is given by a joint policy improvement objective for all intentions. We define the action-value function $Q_\cT(\bs_t, \ba_t)$ for task $\cT$ as 
\begin{equation}
Q_\cT(\bs_t, \ba_t) = r_\cT(\bs_t, \ba_t) + \gamma  \bE_{\pi_\cT} \Big\lbrack R_\cT(\tau_{t+1:\infty}) \Big\rbrack,
\end{equation}
where we have introduced the short-hand notation ${\pi_\cT=\pi_\btheta(\ba | \bx, \cT)}$. Using this definition we define the (joint) policy improvement objective as finding $\arg \max_\btheta \cL(\btheta)$ where $\theta$ is the collection of all intention parameters and,
\begin{alignat}{3}
&\ \ \ &\cL(\btheta) &= \cL(\btheta; \cM) + \sum_{k=1}^{|\sA|} \cL(\btheta; \cA_k), \\
&\text{with}\ \ &\cL(\btheta; \cT) &= \sum_{\cB \in \sT} \mathop{\bE}_{p(s | \cB)} \Big\lbrack Q_\cT(\bs, \ba) \mid \ba \sim \pi_\btheta(\cdot | \bs, \cT)  \Big\rbrack.
\label{eq:obj_pol}
\end{alignat}
That is, we optimize each intention to select the optimal action for its task starting from an initial state drawn from the state distribution $p(\bs | \cB)$, \emph{obtained by following any other policy $\pi(\ba | \bs, \cB)$} with $\cB \in \sT = \cA \cup \lbrace \cM \rbrace$ (the task which we aimed to solve before). We note that this change is a subtle, yet important, departure from a multi-task RL formulation. By training each policy on states sampled according to the state visitation distribution of each possible task we obtain policies that are ``compatible'' -- in the sense that they can solve their task irrespective of the state that the previous intention-policy left the system in. This is crucial if we want to safely combine the learned intentions.

\paragraph{Learning the scheduler}
\label{sub:sched}
The second part of our hierarchical objective is concerned with learning a scheduler that sequences intention-policies. We consider the following setup: Let $\xi$ denote the period at which the scheduler can switch between tasks.\footnote{We choose $\xi = 150$ in our experiments. In general $h$ should span multiple time-steps to enforce commitment to one task.}
Further denote by $H$ the total number of possible task switches within an episode\footnote{We consider a finite horizon setting in the following to simplify the presentation.} and
denote by $\cT_{0:H-1} = \lbrace \cT_0, \dots, \cT_{H-1} \rbrace$ the $H$ scheduling choices made within an episode.
We can define the return of the main task given these scheduling choices as 
\begin{equation}
\begin{aligned}
R_\cM(\cT_{0:H-1}) &= \sum^H_{h = 0} \sum^{(h+1)\xi-1}_{t=h\xi} \gamma^t r_\cM(\bs_t, \ba_t), \\ 
\text{where } \ba_t &\sim \pi_\btheta(\cdot | \bs_{t}, \cT_{h}).
\end{aligned}
\label{eq:sched_ret}
\end{equation}

Denoting the scheduling policy with $P_\cS(\cT | \cT_{0:h-1})$ we can define the probability of an action $\ba_t$, when behaving according to the scheduler, as
\begin{equation}
\pi_\cS(\ba_t | \bs_t, \cT_{0:h-1}) = \sum_\cT \pi_\btheta(\ba_t | \bs_t, \cT) P_\cS(\cT | \cT_{0:h-1}),
\label{eq:sched_policy}
\end{equation}
from which we can sample in two steps (as in Eq. \eqref{eq:sched_ret}) by first choosing a sub-task every $\xi$ steps and then sampling an action from the corresponding intention.
Combining these two definitions, the objective $\cL(\cS)$ for learning a scheduler $\mathcal{S}$ -- by finding the solution $\arg \max_\cS \cL(\cS)$ -- reads:
\begin{equation}
\cL(\cS) = \bE_{P_\cS} \Big\lbrack R_\cM(\cT_{0:H-1}) \mid \cT_h \sim P_\cS(\cT | \cT_{0:h-1}) \Big\rbrack.
\label{eq:obj_sched}
\end{equation}
Note that, for the purpose of optimizing the scheduler, we consider the individual intentions as fixed in Equation \eqref{eq:obj_sched} -- i.e. we do not optimize it w.r.t. $\btheta$ -- since we would otherwise be unable to guarantee preservation of the individual intentions (which are needed to efficiently explore in the first place). We also note that the scheduling policy, as defined above, ignores the dependency on the state $\bs_{h\xi}$ in which a task is scheduled (i.e. $P_\cS$ uses a partially observed state).
In addition to this learned scheduler we also experiment with a version that schedules intentions at random throughout an episode, which we denote with SAC-U in the following. Note that such a strategy is not as naive as it initially appears: due to the fact that we allow several intentions to be scheduled within an episode they will naturally provide curriculum training data for each other. A successful 'move object' intention will, for example, leave the robot arm in a position close to the object, making it easy for a lift intention to discover reward.

As mentioned in Section \ref{sec:related} the problem formulation described above bears similarities to several other multi-task RL formulations. In particular we want to highlight that it can be interpreted as a generalization of the IUA and UNREAL objectives \citep{cabi2017,Jaderberg2017Unreal} to stochastic continuous controls -- in combination with active execution of auxiliary tasks and (potentially learned) scheduling within an episode. It can also be understood as a hierarchical extension of Hindsight Experience Replay \citep{andrychowicz2017hindsight}, where the agent behaves according to a fixed set of semantically grounded auxiliary tasks -- instead of following random goals -- and optimizes over the task selection.

\subsection{Policy Improvement}
\label{sec:policy_improve}
To optimize the objective from Equation \eqref{eq:obj_pol} we take a gradient based approach. We first note that learning for each intention $\pi(\ba | \bs, \cT)$, as defined in Equation \eqref{eq:obj_pol}, necessitates an off-policy treatment -- since we want each policy to learn from data generated by all other policies. 
To establish such a setup we assume access to a parameterized predictor $\hat{Q}^\pi_\cT(\bs, \ba; \phi)$ (with parameters $\phi$) of state-action values; i.e. $\hat{Q}^\pi_\cT(\bs, \ba; \phi) \approx Q^\pi_\cT(\bs, \ba)$ -- as described in Section \ref{sec:policy_eval}.
Using this estimator, and a replay buffer $B$ containing trajectories $\tau$ gathered from all policies, the policy parameters $\btheta$ can be updated by following the gradient
\begin{equation}
\nabla_\btheta \cL(\btheta) \approx \sum_{\substack{\cT \in \sT \\ \tau \sim B}} \mathop{\nabla_{\btheta} \bE}_{\substack{\pi_\btheta(\cdot | \bs_t, \cT) \\ \bs_t \in \tau}} \Big\lbrack \hat{Q}^{\pi}_\cT(\bs_t, \ba; \phi) + \alpha \log \pi_\btheta(\ba | \bs_t, \cT) \Big\rbrack,
\label{eq:policy_improve}
\end{equation}
where $\bE_{\pi_\btheta(\cdot | \bs_t, \cT)} \lbrack \log \pi_\btheta(\ba | \bs_t, \cT) \rbrack$ corresponds to an additional (per time-step) entropy regularization term (with weighting parameter $\alpha$).
This gradient can be computed via the reparametrization trick \citep{rezende14,Kingma2013auto} for policies whose sampling process is differentiable (such as the Gaussian policies used in this work), as described in the work on stochastic value gradients \citep{heess2015learning}. We refer to the supplementary material for a detailed derivation.

In contrast to the intention policies, the scheduler has to quickly adapt to changes in the incoming stream of experience data -- since the intentions change over time and hence the probability that any intention triggers the main task reward is highly varying during the learning process. 
To account for this, we choose a simple parametric form for the scheduler: Assuming a discrete set of tasks $\sT$ we can first realize that the solution 
\begin{equation}
    P_\cS = \arg \max_{P_\cS} \cL(\cS),
    \label{eq:max_sched}
\end{equation} to Equation \eqref{eq:obj_sched} can be approximated by the Boltzmann distribution 
\begin{equation}
P_\cS(\cT_h | \cT_{1:h-1}; \eta) = \frac{\exp(\bE_{P_\cS}[R_\cM(\cT_{h:H})] / \eta)}{\sum_{\bar{\cT}_{h:H}} \exp(\bE_{P_\cS}\lbrack R_\cM(\bar{\cT}_{h:H})\rbrack /\eta)},
\label{eq:softmax_sched}
\end{equation}
where the temperature parameter $\eta$ dictates the greediness of the schedule; and hence $\lim_{\eta \rightarrow \infty} P_\cS(\cT | \cT_{1:h-1}; \eta)$ corresponds to the optimal policy (the solution from \eqref{eq:max_sched}) at any scheduling point. To be precise, the Boltzmann policy corresponds to maximizing $\cL(\cS)$ together with an additional entropy regularizer on the scheduler.

This distribution can be represented via an approximation of the schedule returns $Q(\cT_{1:h-1}, \cT_h) \approx \bE_{P_\cS} [ R_\cM(\cT_{h:H}) | \cT_{1:h-1} ]$.
For a finite, small, number of tasks -- as in this paper -- $Q(\cT_{1:h-1}, \cT_h)$ can be represented in tabular form. Specifically, we form a Monte Carlo estimate of the expectation, using the last $M = 50$ executed trajectories, which yields
\begin{equation}
Q(\cT_{0:h-1}, \cT_h) = \frac{1}{M} \sum_{i=1}^M R^\tau_\cM(\cT_{h:H}),
\end{equation}
where $R^\tau_\cM(\cT_{h:H})$ is the cumulative discounted return along trajectory $\tau$ (computed as in Equation \eqref{eq:sched_ret} but with fixed states and action choices).

Using the improved policy from Equation \eqref{eq:policy_improve} and the scheduler defined via Equation \eqref{eq:softmax_sched} we can then collect addition data by following the scheduled action distribution $\pi_\cS(\ba | \bs, \cT_{q:h-1})$ given by Equation \eqref{eq:sched_policy}.

\subsection{Policy Evaluation}
\label{sec:policy_eval}
We use Retrace \citep{Munos16} for off-policy evaluation of all intentions. Concretely, we train parametric Q-functions (neural networks) $\hat{Q}^\pi_\cT(\bs, \ba; \phi)$ by minimizing the following loss, defined on data from the replay $B$:
\begin{equation}
\begin{aligned}
  &\min_\phi L(\phi) = \bE_{(\tau, b, \cB) \sim B} \Big[ \big( \hat{Q}^\pi_\cT(\bs, \ba; \phi) - Q^{\text{ret}} \big)^2 \Big], \text{with } \\ 
  &Q^{\text{ret}} = \sum_{j=i}^\infty \Big( \gamma^{j-i} \prod_{k=i}^j c_k \Big) \Big[ r_\cT(s_j, a_j) + \delta_Q(\bs_i, \bs_j) \Big], \\
  &\delta_Q(\bs_i, \bs_j) = \bE_{\pi_{\theta'}(\ba | \bs, \cT)} [ Q^\pi_\cT(\bs_i, \cdot; \phi') ] - Q^\pi_\cT(\bs_j, \ba_j; \phi'), \\
  &c_k = \min\Big(1, \frac{\pi_{\theta'}(\ba_k | \bs_k, \cT)}{b(\ba_k | \bs_k, \cB)}\Big),
\end{aligned}
\end{equation}
where $\tau$ denotes a trajectory (together with action choices and rewards) sampled from the replay buffer, $b$ denotes a behaviour policy under which the data was generated, and $\cB$ is the task the behaviour policy tried to accomplish. We again highlight that $b$ was not necessarily aiming to achieve task $\cT$ for which $\hat{Q}_\cT$ should predict action-values. The importance weights $c_k$ then weight the actions selected under the behavior policy with their probability under $\pi$. Here $\phi'$ and $\theta'$ denote the parameters of target policy and Q-networks \citep{Mnih15}, which are periodically exchanged with the current parameters $\phi, \theta$. This is common practice in Deep-RL algorithms to improve learning stability.

\section{Experiments}

To benchmark our method we perform experiments based on a Kinova Jaco robot arm in simulation and on hardware.

\subsection{Experimental Setup}

In all experiments the auxiliary tasks are chosen to provide the agent with information about how well it is exploring its own sensory space. They are easy to compute and are general -- in the sense that they transfer across tasks. They are defined over all available sensor modalities. For proprioception, for example, we choose to maximize / minimize joint angles, for the haptic sensors we define tasks for activating / deactivating finger touch or force-torque sensors. In image space, we define auxiliary tasks on the object level (i.e. 'move red object' or 'place red object close to green object in camera plane'). All these predicates can be easily computed and mapped to a sparse reward signal (as in Equation \eqref{eq:sparse_rew}).
A full list of auxiliary rewards can be found in the supplementary material.

We present learning results for SAC-X with the two schedulers described in Section \ref{sub:sched}: a sequentially uniform scheduler SAC-U and SAC-X with a learned scheduler SAC-Q. In ablation studies, we also investigate a non-scheduling
version of our setup, where we strictly followed the policy that optimizes the external reward. Since this procedure is similar to the one used by the IU agent \citep{cabi2017} -- but enhanced with retrace and stochastic policies to ensure an even comparison --, we denote this variant with 'IUA' in the following. As a strong off-policy learning baseline we also include a comparison to DDPG \citep{Lillicrap16}.

All simulation experiments use raw joint velocities (9 DOF) as control signals at 50~ms time steps. Episodes lasted for 360 time-steps in total with scheduler choices every $\xi= 180$ steps. Observations consist of proprioceptive information (joint angles, joint velocities) of the arm as well as sensor information coming from a virtual force-torque sensor in the wrist, virtual finger touch sensors and simulated camera images. We provide results for both learning from raw pixels and learning from extracted image features (i.e. information about pose and velocities of the objects contained in the scene) we refer to the supplementary material for details on the policy network architecture.
All experiments are repeated with 5 different random seeds; learning curves report the median performance among the 5 runs (with shaded areas marking the $5 \%$ and $95 \%$ quantiles respectively).

To speed up experimentation, all simulation results are obtained in an off-policy learning setup where data is gathered by \textbf{multiple agents (36 actors)} which send collected experience to a pool of learners (36 learners were used). This setup is explained in more detail in the Supplementary material. While this is a compromise on data-efficiency -- trading it off with wall-clock time -- our real world experiments, in which a single robot is the only data source, reveal that SAC-X can be very data-efficient.

\begin{figure}[htbp]
\begin{center}
\centerline{\includegraphics[width=\columnwidth]{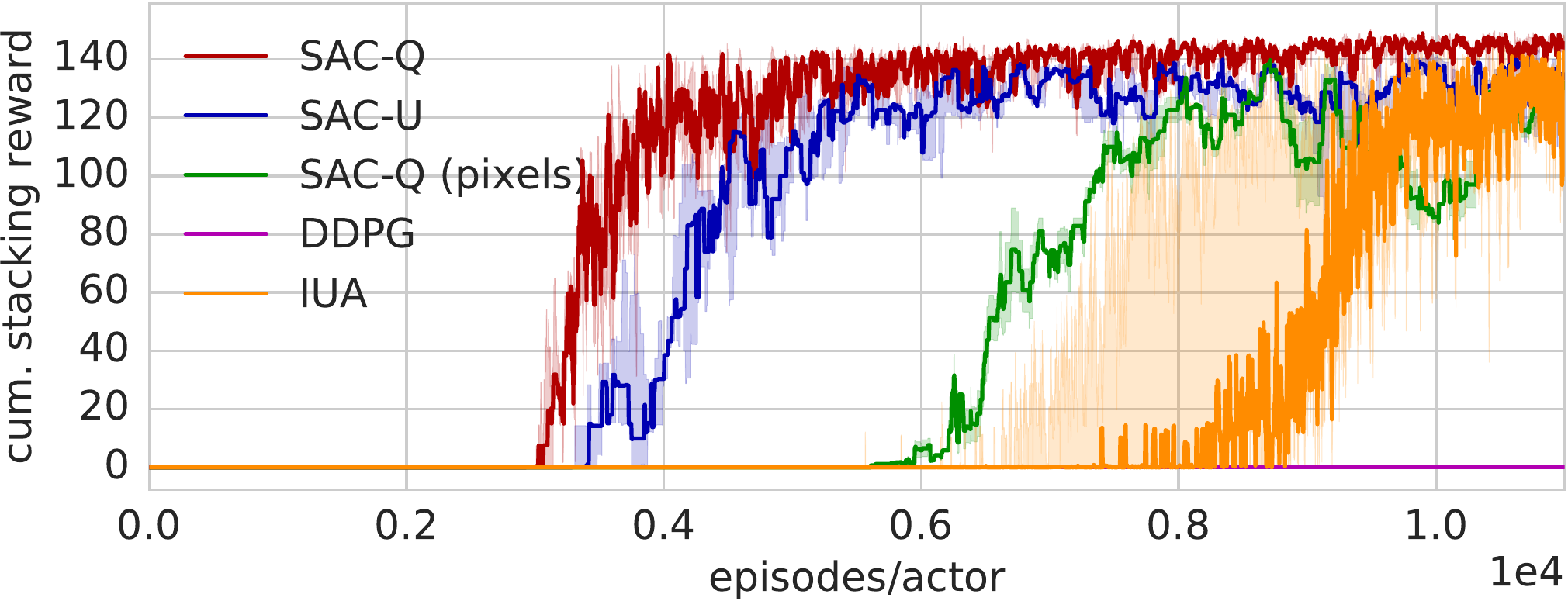}}
\caption{Cumulative reward for the extrinsic task of stacking block one on block two.
Both SAC-U and SAC-Q learn the task reliably. The reference experiment using DDPG fails
completely (flat line). The IUA approach learns slower and less reliably. Note that all results were obtained via 36 actors and learners.}
\label{stack_1-extrinsic-plot}
\end{center}
\vskip -0.4in
\end{figure}
\subsection{Stacking Two Blocks}
\label{sect:stack_exp}
\begin{figure}[t]
\begin{center}
\includegraphics[width=0.9\columnwidth]{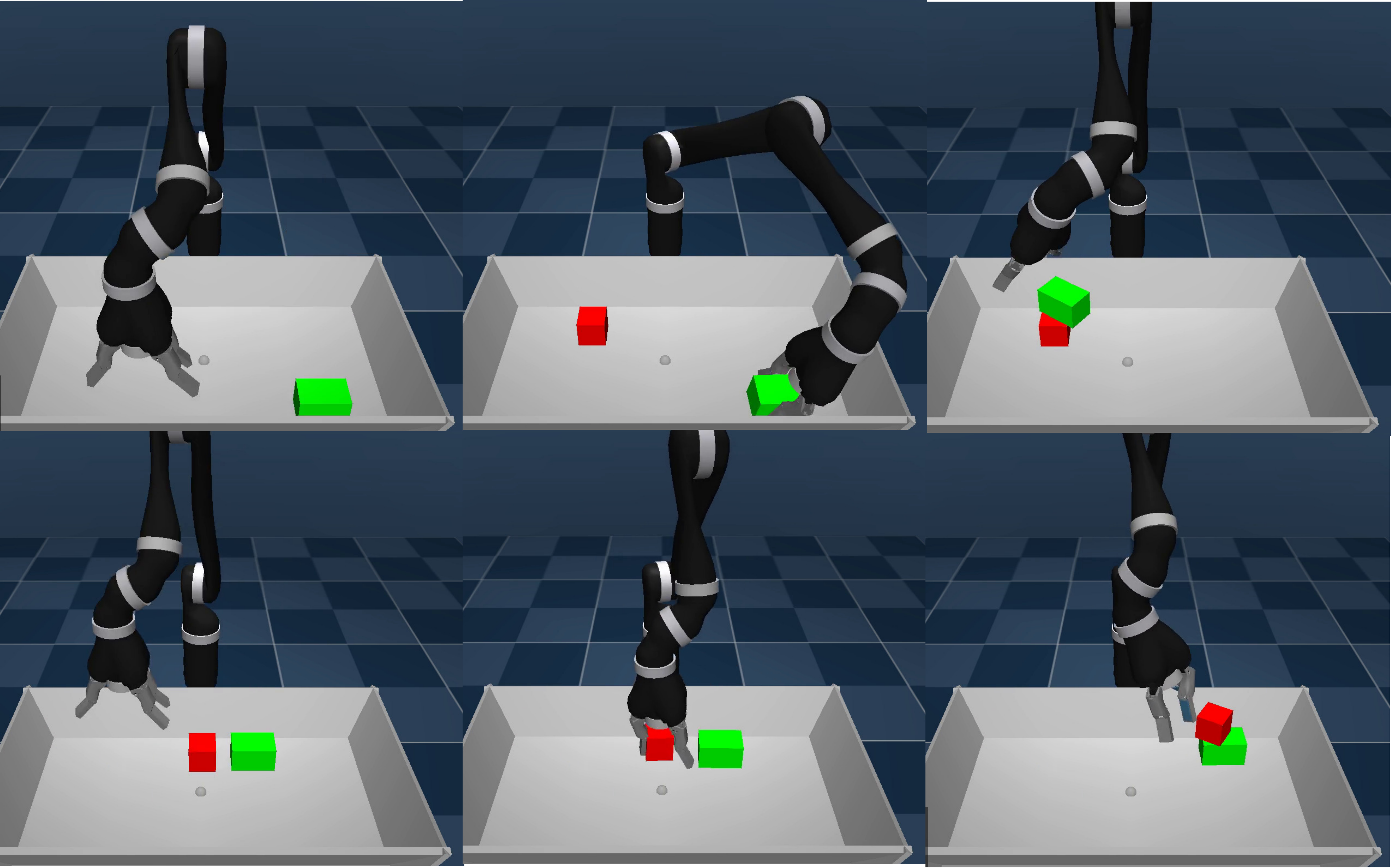}
\caption{Depiction of the agent stacking two blocks in either configuration, red above green or vice-versa.}
\label{fig:stack2-sequence}
\end{center}
\vskip -0.2in
\end{figure}

For our initial set of algorithm comparisons we consider the task of stacking a block on top of another, slightly larger, object. This constitutes a challenging robotics task as it requires the agent to acquire several core abilities:
grasping the first block placed arbitrarily in the workspace, lifting it up to a certain height, precisely placing it on top of the second block. In addition, the agent has to find a stable configuration of the two blocks. The expected behavior is shown in the bottom image sequence in Figure \ref{fig:stack2-sequence}.
We use a sparse reward for a successful stack: the stack reward is one if the smaller object is only in contact with other objects in the scene, but not with the robot or the ground. Otherwise the reward is zero.
In addition to this main task reward the agent has access to the standard set of auxiliary rewards, as defined in the supplementary material.

Figure \ref{stack_1-extrinsic-plot} shows a comparison between SAC-X and several baselines in terms of of average stacking reward. As shown in the plot, both SAC-U (uniform scheduling) and {SAC-Q} reliably learn the task for all seeds. SAC-U reaches a good performance after around 5000 episodes per actor, while SAC-Q is faster and achieves a slightly better final performance -- thanks to its learned scheduler. To demonstrate that our method is powerful enough to learn policies and action-value functions from raw images, we performed the same stacking experiment with information of the block positions replaced by two camera images of the scene -- that are processed by a CNN and then concatenated to the proprioceptive sensor information (see supplementary material for details). The results of this experiment reveal that while learning from pixels (SAC-Q (pixels)) is slower than from features, the same overall behaviour can be learned.

\begin{figure}[t]
\begin{center}
\centerline{\includegraphics[width=\columnwidth]{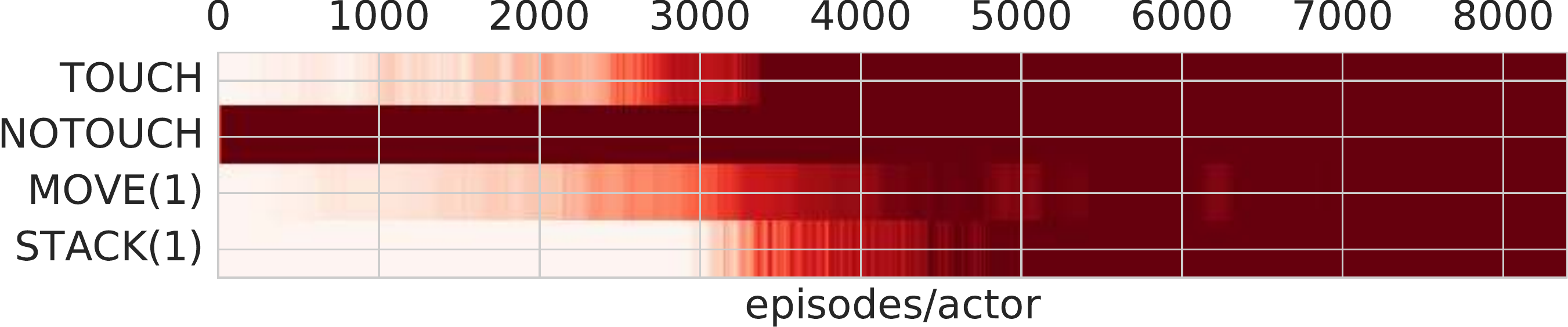}}
\caption{Learning times for a subset of the 13 auxiliary intentions we used in the SAC-Q approach and for the external stacking task. Red color codes for reward. First the agent learns to interact with the objects by touching them and moving them around, then more complex intentions can be learned until, finally, stacking is learned.}
\label{fig:stack_1-sac-auxplot}
\end{center}
\vskip -0.2in
\end{figure}

In the no scheduling case, i.e. when the agent follows its behaviour policy induced by the external reward ('IUA'), the figure reveals occasional successes in the first half of the experiment, followed by late learning of the task. Presumably learning is still possible since the shared layers in the policy network bias behaviour towards touching/lifting the brick (and Retrace propagates rewards along trajectories quickly, once observed). But the variability in the learning process is much higher and learning is significantly slower. Finally, DDPG fails on this task; the reason being that a stacking reward is extremely unlikely to be observed by pure random exploration and therefore DDPG can not gather the data required for learning. Both results support the core conjecture: scheduling and execution of auxiliary intentions enables reliable and successful learning in sparse reward settings. 
Figure \ref{fig:stack_1-sac-auxplot} gives some insight into the learning behaviour, plotting a subset of the learned intentions (see the supplementary for all results). The agent first learns to touch (TOUCH) or stay away from the block (NOTOUCH) then it learns to move the block and finally stack it.

\subsection{Stacking a 'Banana' on Top of a Block}

Using less uniform objects than simple blocks poses additional challenges, both for grasping and for stacking: some object shapes only allow for specific grasps or are harder to stack in a stable configuration. We thus perform a second experiment in which a banana shaped object must be placed on top of a block. For an approach relying on shaping rewards, this would require careful re-tuning of the shaping. With SAC-X, we can use the same set of auxiliary tasks.

\begin{figure}[tbp]
\begin{center}
\centerline{\includegraphics[width=\columnwidth]{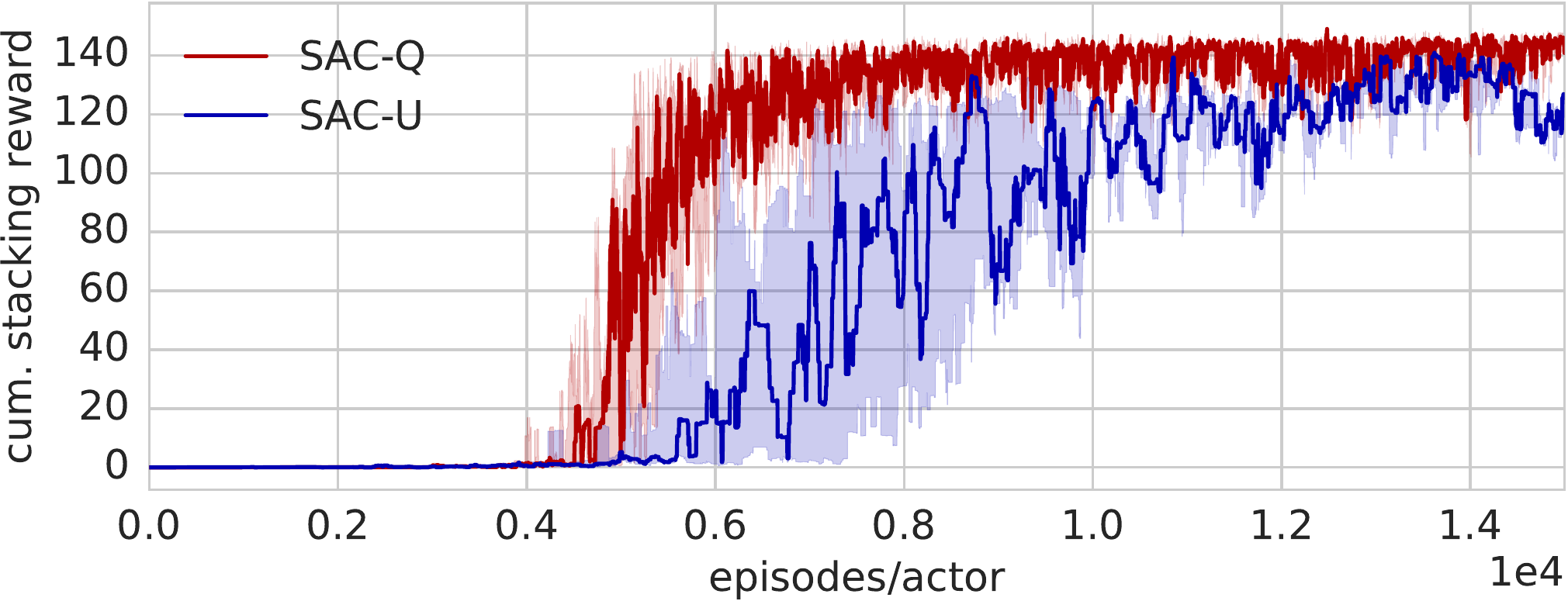}}
\caption{Comparison, in terms of cumulative reward, between SAC-Q and SAC-U for the 'banana' stacking experiment.}
\label{fig:stack_banana}
\end{center}
\vskip -0.3in
\end{figure}

Figure \ref{fig:stack_banana} depicts the results of this experiment. Both SAC-U and SAC-Q can solve the task. In this case however, the advantages of a learning scheduler that focuses on solving the external task become more apparent. One explanation for this is that stacking the banana does require a careful fine-tuning of the stacking policy -- on which the learned scheduler naturally focuses. 

\begin{figure}[h]
\begin{center}
\centerline{\includegraphics[width=\columnwidth]{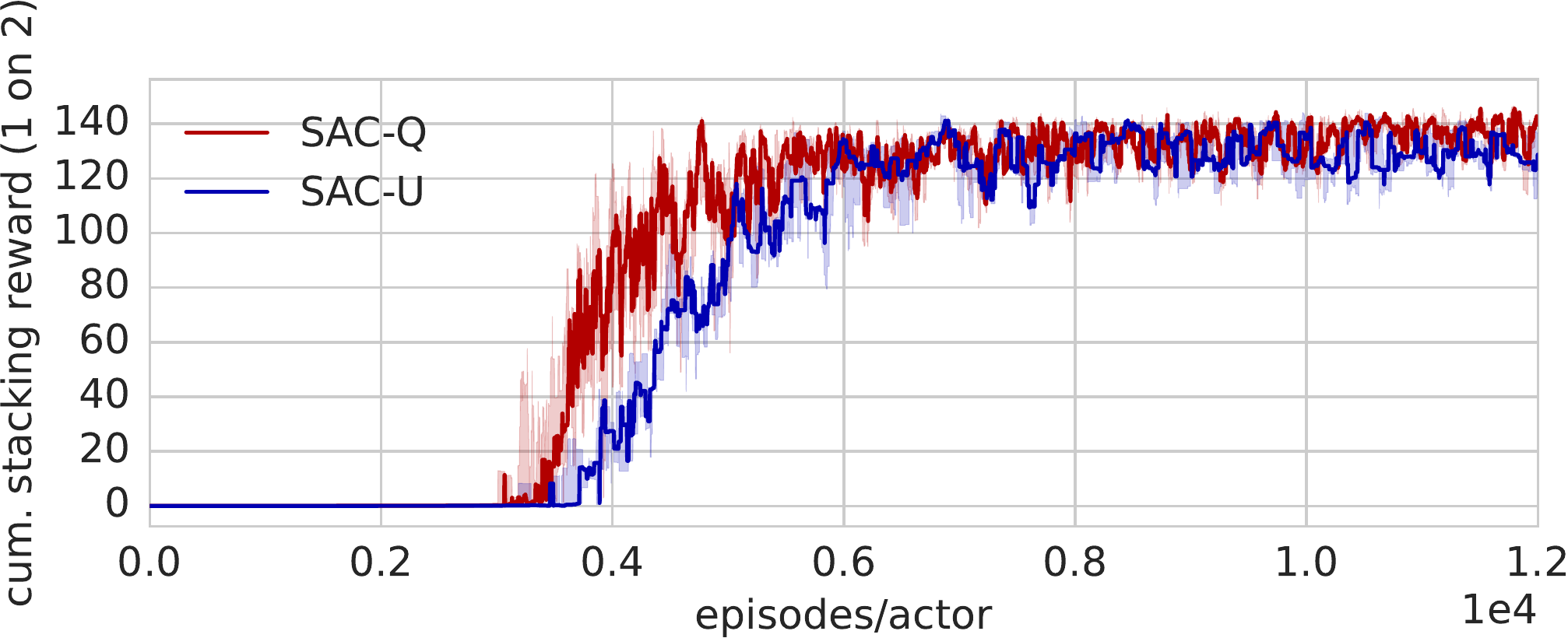}}
\centerline{\includegraphics[width=\columnwidth]{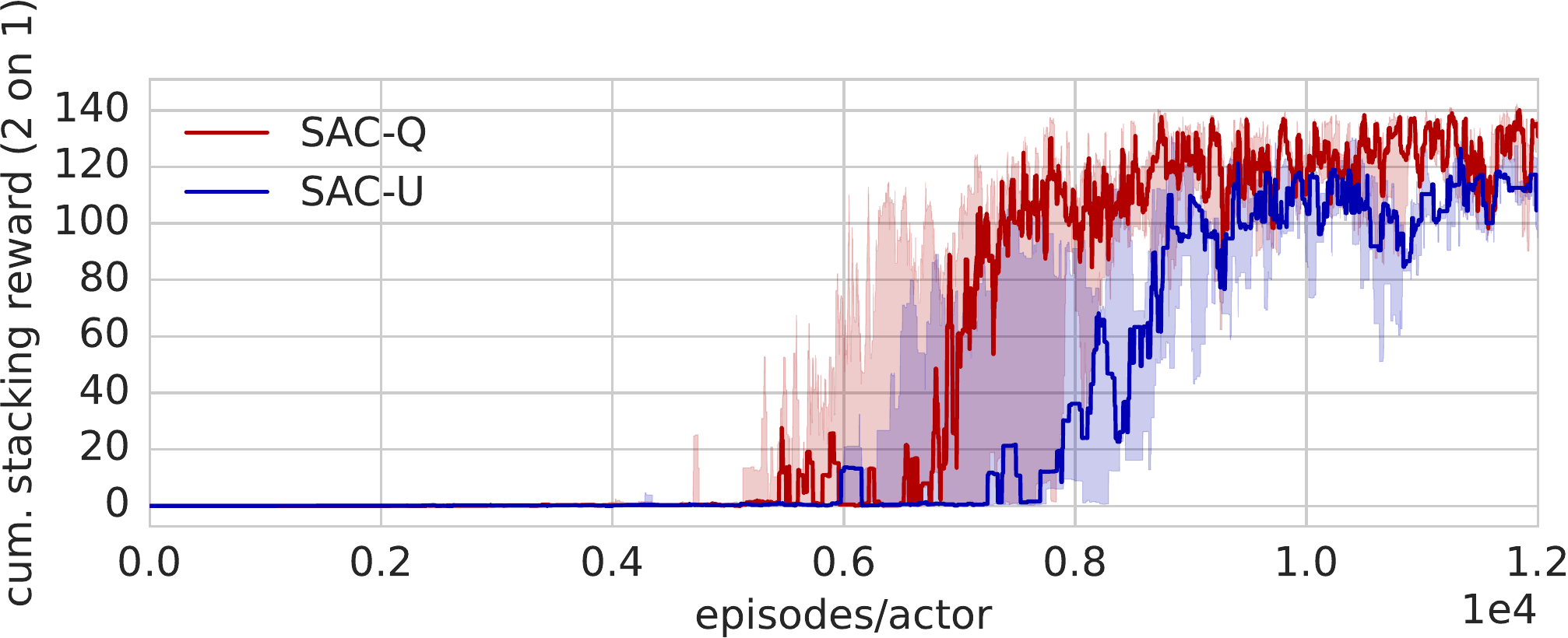}}
\caption{Comparison, in terms of learning speed, between SAC-Q and SAC-U for the two block stacking task.}
\label{fig:stack-2}
\end{center}
\vskip -0.2in
\end{figure}

\subsection{Stacking Blocks Both Ways}

Next we extend the stacking task by requiring the agent to both: stack the small red block on the large green block (1 on 2 in the Figure \ref{fig:stack2-sequence}) as well as vice-versa (2 on 1 in the figure). This is an example of an agent learning multiple external tasks at once. To cope with multiple external tasks, we learn multiple schedulers (one per task) and pick between them at random (assuming external tasks have equal importance).

Both SAC-U and SAC-Q are able to accomplish the external tasks from pure rewards (see Figure \ref{fig:stack-2}). As is also apparent from the figure, the SAC-X agent makes efficient use of its replay buffer: Compared to the initial stacking experiment (Section \ref{sect:stack_exp}), which required 5000 episodes per actor, SAC-X only requires 2500 additional episodes per actor to learn the additional task.
In addition to this quantitative evaluation, we note that the observed behaviour of the learned agent also exhibits intuitive strategies to deal with complicated situations. For example, if the agent is started in a situation where block one is already stacked on block two, it has learned to first put block one back on the table, and then stack block two on top of the first block - all in one single policy (please also see the supplementary \href{https://youtu.be/mPKyvocNe_M}{video} at 0:50 mins for a demonstration).

\begin{figure}[t]
\vskip 0.1in
\begin{center}
\centerline{
 \includegraphics[width=\columnwidth]{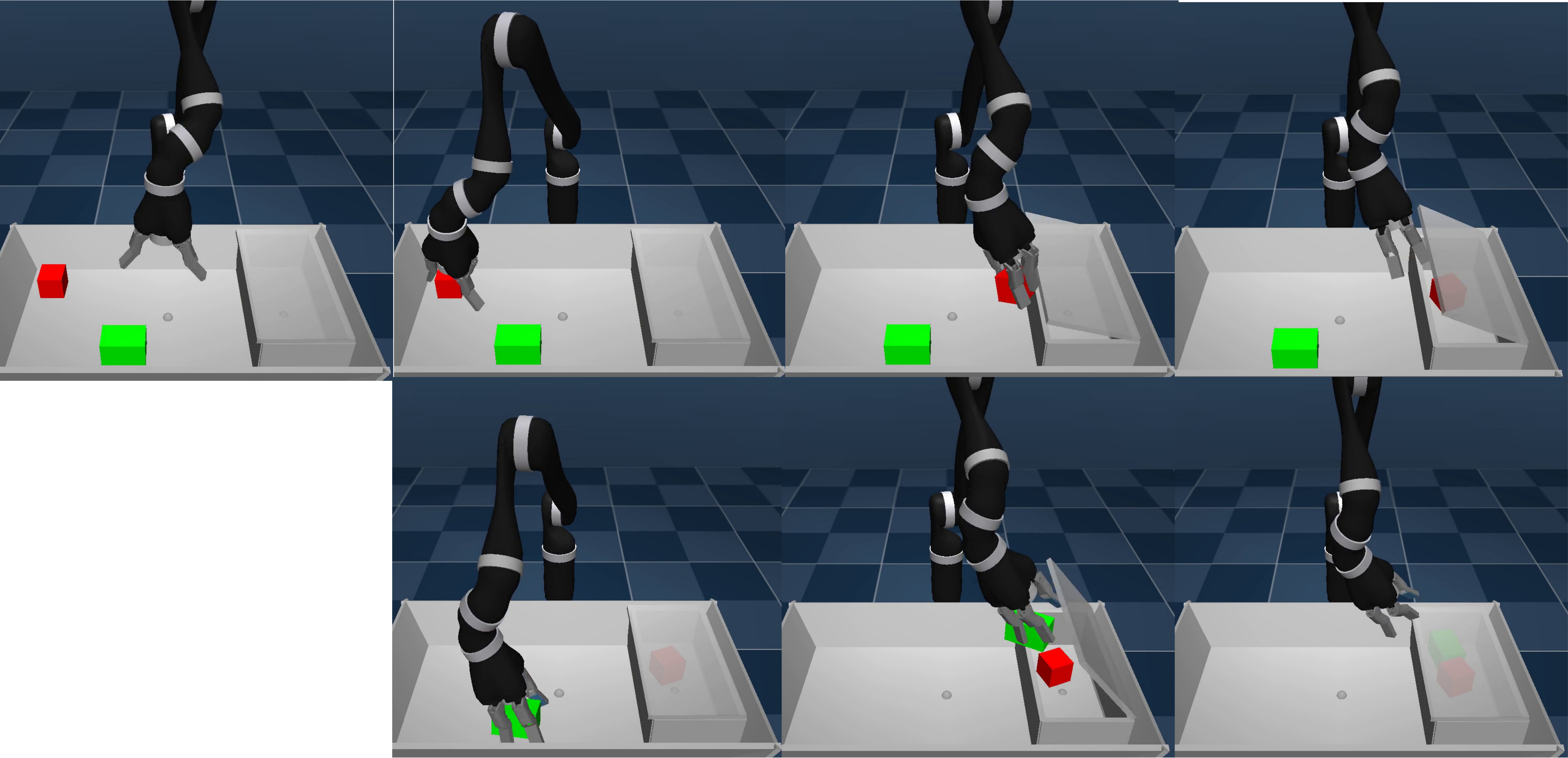}}
\caption{The 'clean-up' task. The images depict a trajectory (left-to-right, top-to-bottom) of the final behaviour for the 'put all in box' intention.}
\label{cleanup-sequence}
\end{center}
\vskip -0.3in
\end{figure}

\begin{figure}[t]
\vskip 0.1in
\begin{center}
\centerline{\includegraphics[width=\columnwidth]{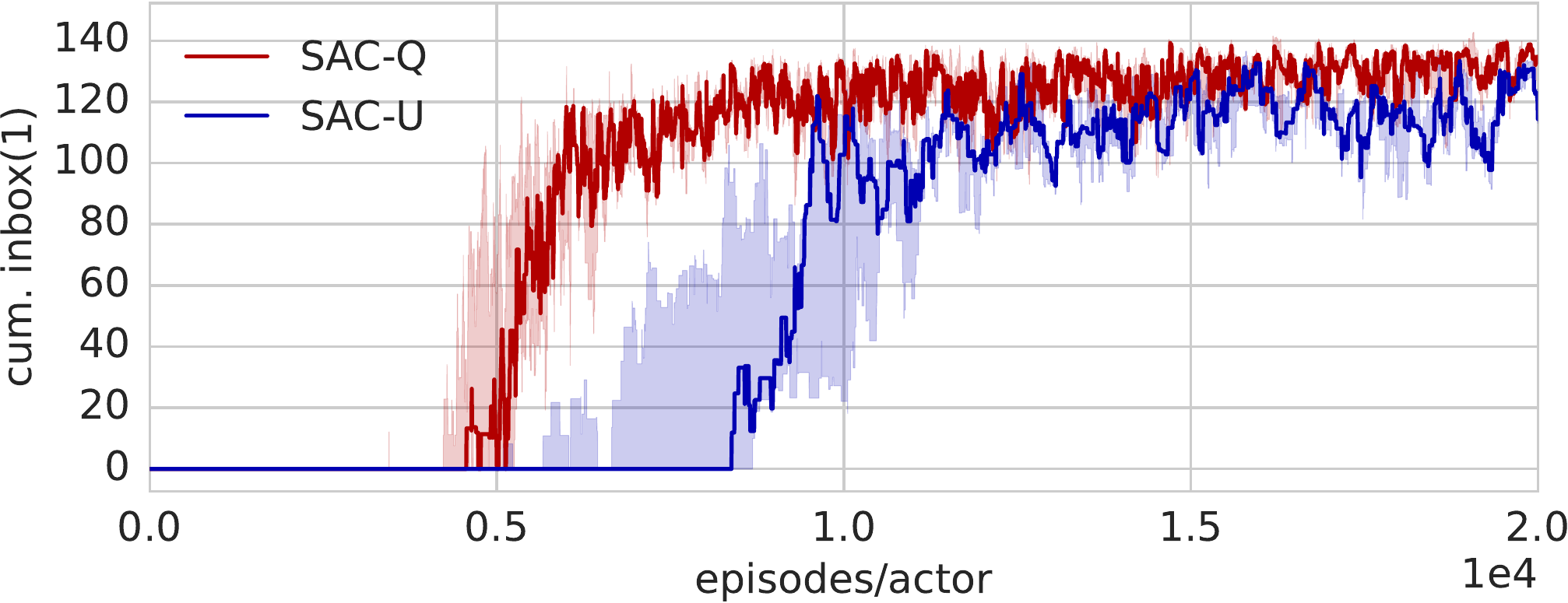}}
\centerline{\includegraphics[width=\columnwidth]{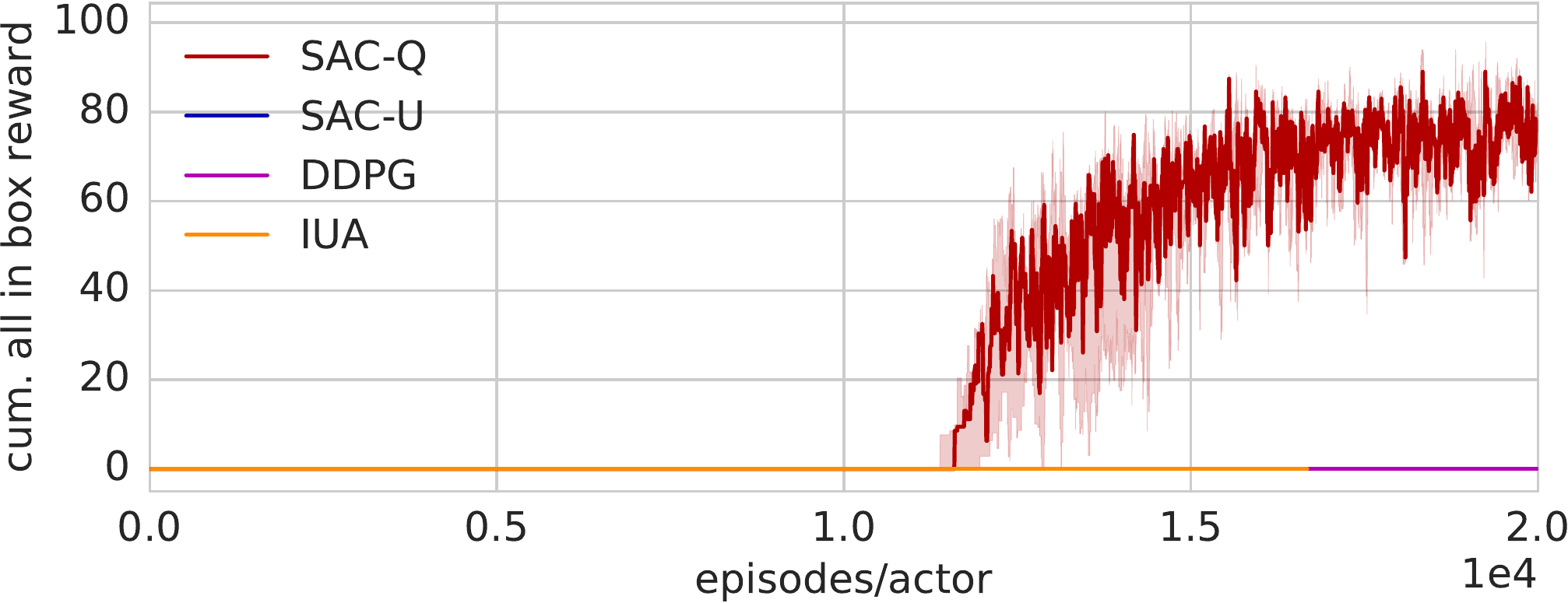}}
\caption{Learning for the cleanup task, shown is the most difficult external task where two blocks are required to be in the box to get a reward signal. SAC-Q is the only successful approach (bottom). SAC-U here 'only' learns to put a single block into the box (top).}
\label{cleanup-sac}
\end{center}
\vskip -0.3in
\end{figure}

\subsection{The 'clean-up' Task}
The clean-up task (see Figure \ref{cleanup-sequence}) is an example where a sequence of specific movements have to be executed in order to solve the task.
In addition to the two different sized blocks from the last experiments, we add a new object to the scene: a static box with a lid that can be opened.

\begin{figure}[t]
\begin{center}
\centerline{\includegraphics[width=\columnwidth]{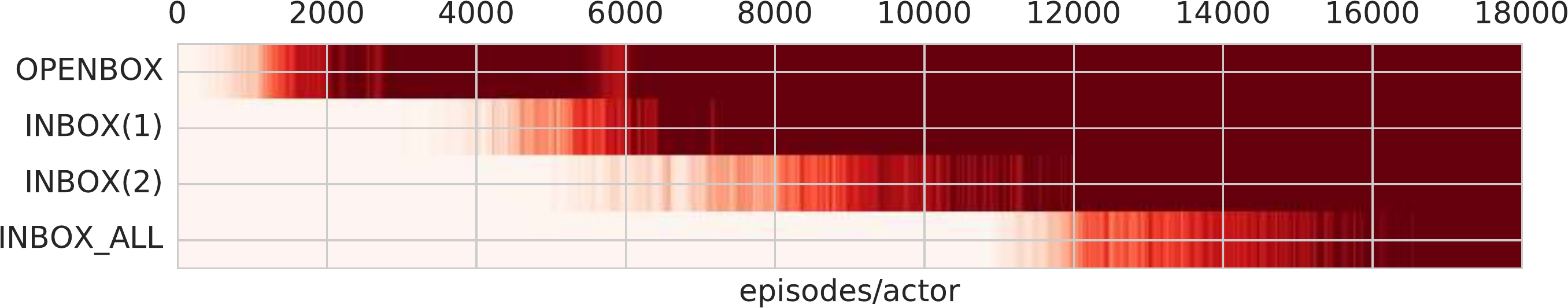}}
\caption{Expected reward in the 'clean-up' experiment, SAC-Q learns alll four extrinsic tasks reliably. In addition it reliably learns to also solve the 15 auxiliary tasks (not depicted here).}
\label{cleanup-sac-alt}
\end{center}
\vskip -0.4in
\end{figure}

We rely on the same auxiliary tasks as in the stack blocks experiment, adding one additional sparse auxiliary intention for each object in relation to the box: 'bring object above and close to the box'.
In contrast to previous experiments, we now have 4 sparse extrinsic tasks and corresponding intention policies: i) open the box (OPENBOX in the Figure), ii) put object 1 in box (INBOX(1)), iii) put object 2 in box (INBOX(2)), and iv) put all objects in the box (INBOX\_ALL).
With a total of 15 auxiliary and 4 extrinsic tasks, this is the most complex scenario presented in this paper. Figure \ref{cleanup-sac} shows a comparison between SAC-X and baselines for this task. Remarkably, even though the reward for placing the objects into the box can \emph{only be observed once they are correctly placed}, SAC-Q learns all extrinsic tasks (see also Figure \ref{cleanup-sac-alt} and the supplementary for a detailed comparison to SAC-U), and the auxiliary tasks, reliably and can interpolate between intention policies (see supplementary \href{https://youtu.be/mPKyvocNe_M}{video}). All baselines fail in this setting, indicating that SAC-X is a significant step forward for sparse reward RL.

\subsection{Learning from Scratch on the Real Robot}

For learning on the real robot, we consider two tasks: lifting a block and a bring task.
We first checked the feasibility of both tasks in simulation by learning using a single actor run in real-time.
Using SAC-X, both tasks can be successfully learned from pure rewards with full 9 DOF raw joint velocity control.
The learning time on the real robot however would have been the equivalent of several days of non-stop experimentation on the real robot.
For practical feasibility we therefore made the following adaptations:
we used a cartesian controller for velocity based control of the
hand plus one control action for actuation of two fingers, resulting in a 4 dimensional continuous
control vector. Note that the proprioceptive information provided to the controller still consist of the joint positions and velocities.

\begin{figure}[t]
\begin{center}
 \begin{flushright}
    \includegraphics[width=\columnwidth]{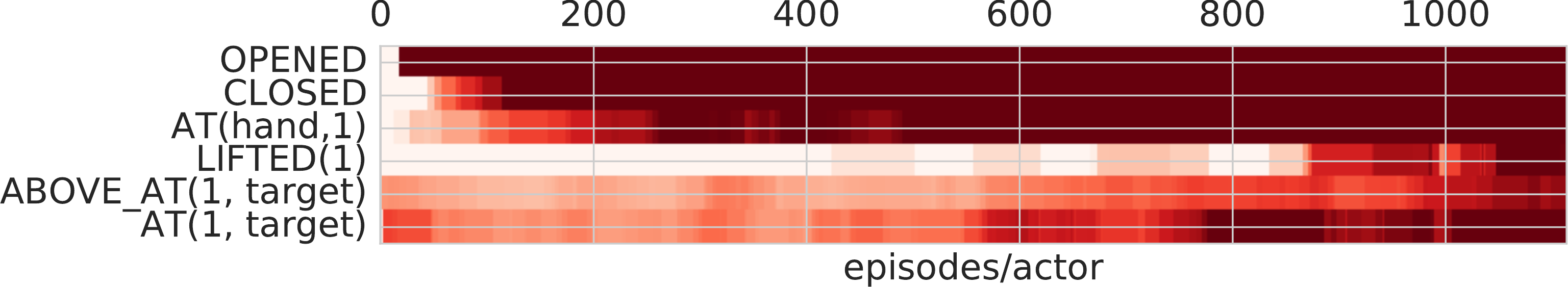} \\
    \includegraphics[width=0.89\columnwidth]{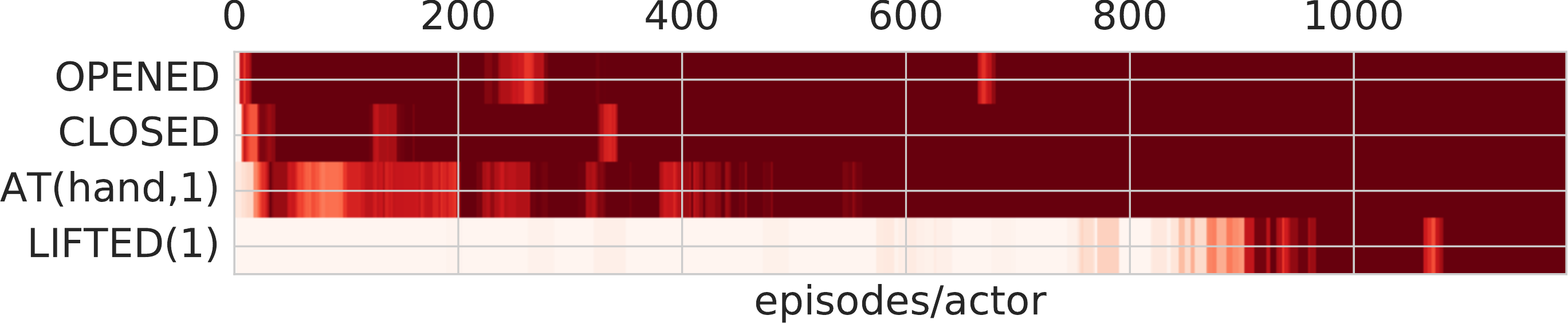}
 \end{flushright}
 \vspace{-0.1in}
\caption{Learning statistics for a real robot experiment for the bring (top) and lift (bottom) task. As before, red indicates reward within an episode (averaged over the last 10), and we plot successes for all used auxiliary tasks (see the supplementary for a detailed listing).}
\label{fig:real_robot_stat}
\end{center}
\vskip -0.2in
\end{figure}

In the lift experiment three auxiliary rewards were defined (rewarding the robot for closing fingers, opening fingers and proximity to the brick).
The learning curves, depicted in Figure \ref{fig:real_robot_stat} (top), reveal that using a single robot arm SAC-Q successfully learns
to lift after about 1200 episodes, requiring about 10 hours of learning time on the real robot.
When tested on about 50 trials on the real robot, the agent is 100\% successful in achieving the lifting task.

In an even more challenging setup, we trained SAC-Q to also place the 
block at in a given set of locations in its workspace; adding additional tasks that reward the agent for reaching said location.
Again, learning was successful (see Figure \ref{fig:real_robot_stat}, bottom), and the agent showed robust, non-trivial control behavior: The resulting policy developed various techniques for achieving the task including dragging and pushing the block with one finger as well as lifting and carrying the block to the goal location. Furthermore, the agent learned to correct the block position of imprecisely placed objects and learned to move the gripper away once the task is completed. This reactive and rich control behaviour is due to the closed-loop formulation of our approach.

\begin{figure}[t]
\vskip 0.2in
\begin{center}
\includegraphics[width=\columnwidth]{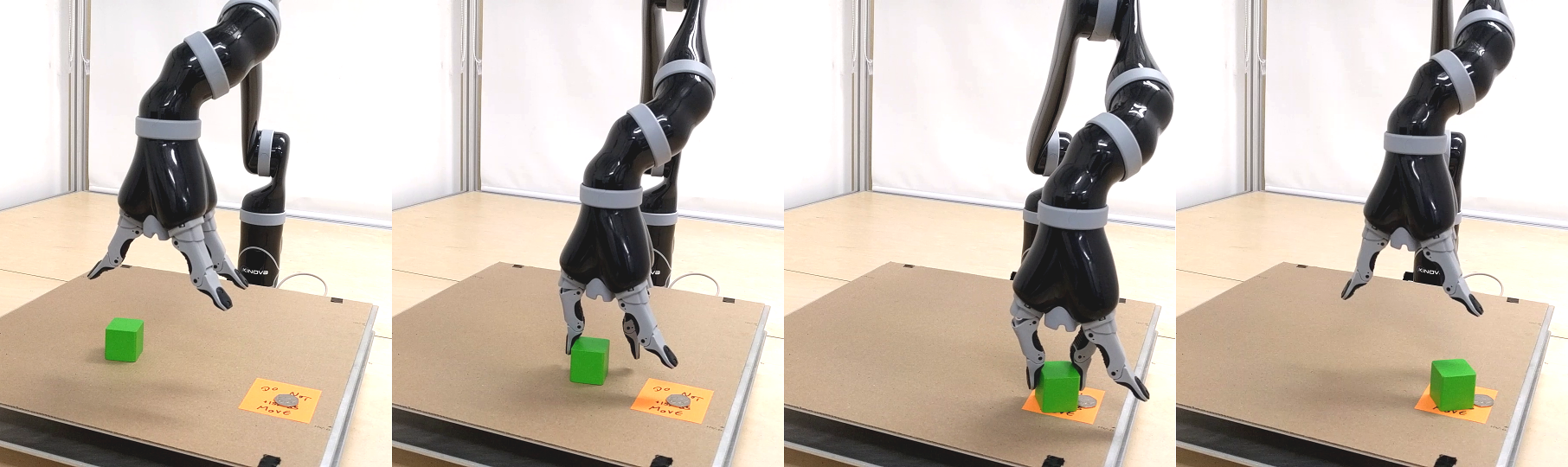} \\
\includegraphics[width=\columnwidth]{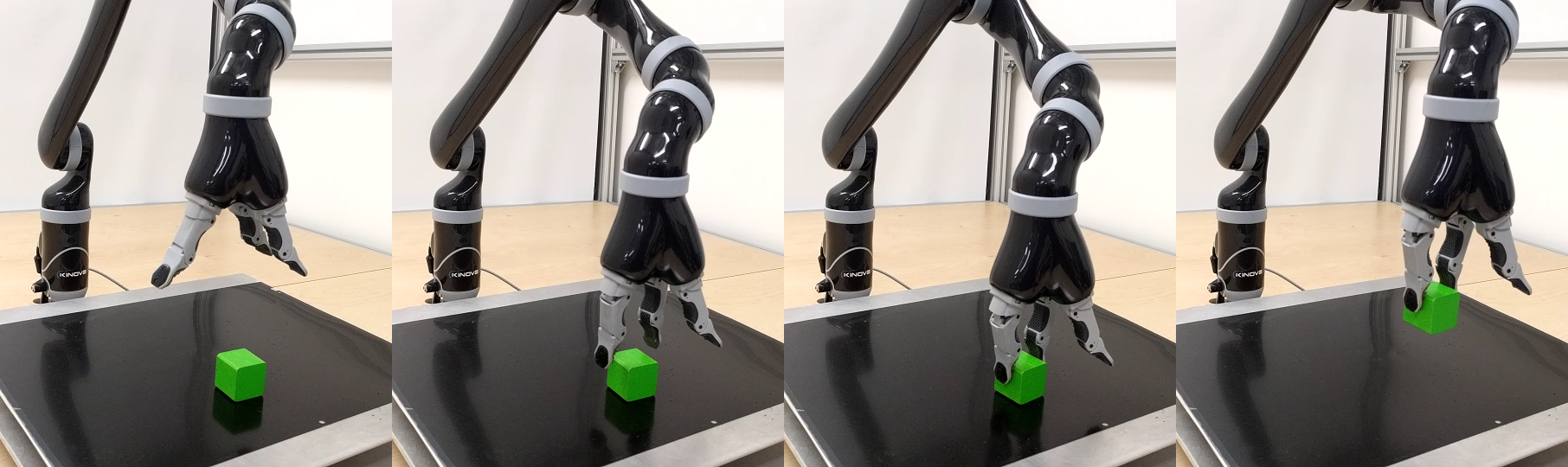}
\vspace{-0.2in}
\caption{Image sequence depicting a trained SAC-Q agent on the real robot solving the bring (top) and lift (bottom) task.}
\label{fig:real_robot_seq_bring}
\end{center}
\vskip -0.2in
\end{figure}

\section{Conclusion}
This paper introduces SAC-X, a method that simultaneously learns intention policies
on a set of auxiliary tasks, and actively schedules and executes these
to explore its observation space - in search for sparse rewards of externally defined target tasks. Utilizing simple auxiliary tasks enables SAC-X to learn complicated target tasks from rewards defined in a 'pure', sparse, manner: only the end goal is specified, but not the solution path.

We demonstrated the power of SAC-X on several challenging
robotics tasks in simulation, using a common set of simple and sparse auxiliary tasks and on a real robot.
The learned intentions are highly reactive, reliable, and exhibit  a rich and robust behaviour. 
We consider this as an important step towards the goal of applying RL to real world domains.

\section{Acknowledgements}
The authors would like to thank Yuval Tassa, Tom Erez, Jonas Buchli, Dan Belov and many others of the DeepMind team for their help and numerous useful discussions and feedback throughout the preparation of this manuscript.

{\small \bibliography{references}}
\bibliographystyle{icml2018}

\newpage
\appendix

\section{Details on the Experimental Setup}

\subsection{Simulation}

For the simulation of the Jaco robot arm the numerical simulator MuJoCo \footnote{MuJoCo: see www.mujoco.org} was used -- using a model we identified from our real robot setup.

The simulation was run with a numerical time step of 10 milliseconds, integrating 5 steps, to get a control interval of 50 milliseconds for the agent. In this way we can resolve all important properties of the robot arm and the object interactions in simulation.

The objects that are used are based on wooden toy blocks.
We use a cubic block with side lengths of 5 cm (red object) and a cuboid with side lengths of 5cm x 5cm x 8cm (green block). For the banana stacking experiment a combination of 3 different geometric (capsule shaped) primitives with radius 2.5 cm are used, resulting in a banana shaped object of 12 cm in length (replacing the red object).

All experiments made use of an experiment table with sides of 60 cm x 30 cm in length, which is assumed to be the full working space for all experiments.
The objects are spawned at random on the table surface. The robot hand is initialized randomly above the table-top with a height offset of up to 20 cm above the table (minimum 10 cm) and the fingers in an open configuration.
The simulated Jaco is controlled by raw joint velocity commands (up to 0.8 radians per second) in all 9 joints (6 arm joints and 3 finger joints).
All experiments run on episodes with 360 steps length (which gives a total simulated real time of 18 seconds per episode).
For the SAC-X experiments we schedule 2 intentions each episode, holding the executed intention fixed for 180 steps.

\begin{table}[t]
\caption{Proprioceptive observations used in all simulation experiments.}
\label{obs-proprio-table}
\vskip 0.15in
\begin{center}
\begin{small}
\begin{tabular}{lcc}
\toprule
Entry & dimensions & unit \\
\midrule
arm joint pos & 6 & rad\\
arm joint vel & 6 & rad / s\\
finger joint pos & 3 & rad\\
finger joint vel & 3 & rad / s\\
finger touch & 3 & N\\
TCP pos & 3 & m\\
\bottomrule
\end{tabular}
\end{small}
\end{center}
\vskip -0.1in
\end{table}

\begin{table}[t]
\caption{Object feature observations, used in the default simulation experiments. For the pixel experiments these observations are not used.
The pose of the objects is represented as world coordinate position and quaternions. In the table m denotes meters, q refers to a quaternion which is in arbitrary units (au).}
\label{obs-features-table}
\vskip 0.15in
\begin{center}
\begin{small}
\begin{tabular}{lcc}
\toprule
Entry & dimensions & unit \\
\midrule
object i pose & 7 & m au\\
object i velocity & 6 & m/s, dq/dt \\
object i relative pos & 3 & m\\
\bottomrule
\end{tabular}
\end{small}
\end{center}
\vskip -0.1in
\end{table}

For the feature based experiments in simulation we make use of the proprioceptive features that the Jaco robot can deliver (see Table \ref{obs-proprio-table}).
In addition, for the default simulation experiments, we use features from the objects in the scene, that are computed directly in simulation (see table \ref{obs-features-table}).
This gives a total of 56 observation entries.
For the cleanup experiment, we add the lid angle and lid angle velocity, which gives a total of 58 observations for this experiment.
For the pixel experiments, we use two RGB cameras with an resolution of 48 x 48 (see table \ref{obs-pixel-table}) in combination with the proprioceptive features (table \ref{obs-proprio-table}).

\begin{table}[t]
\caption{Pixel observations that replace the object observations of table \ref{obs-features-table} for the pixel experiments.}
\label{obs-pixel-table}
\vskip 0.15in
\begin{center}
\begin{small}
\begin{tabular}{lcc}
\toprule
Entry & dimensions & unit \\
\midrule
camera 1 & 48 x 48 x 3 & rgb\\
camera 2 & 48 x 48 x 3 & rgb\\
\bottomrule
\end{tabular}
\end{small}
\end{center}
\vskip -0.1in
\end{table}

\subsubsection{Auxiliary Reward Overview}
\label{sect:aux_rewards}

We use a basic set of general auxiliary tasks for our experiments.
Dependent on the type and number of objects in the scene the number of available auxiliary tasks can vary.

\begin{itemize}
    \item \textit{TOUCH, NOTOUCH}: Maximizing or minimizing the sum of touch sensor readings on the three fingers of the Jaco hand. (see Eq. \ref{eq:sparse_rew_TOUCH} and Eq. \ref{eq:sparse_rew_NOTOUCH})
    \item \textit{MOVE(i)}: Maximizing the translation velocity sensor reading of an object. (see Eq. \ref{eq:sparse_rew_MOVE})
    \item \textit{CLOSE(i,j)}: distance between two objects is smaller than 10cm (see Eq. \ref{eq:sparse_rew_CLOSE})
    \item \textit{ABOVE(i,j)}: all points of object i are above all points of object j in an axis normal to the table plane (see Eq. \ref{eq:sparse_rew_ABOVE})
    \item \textit{BELOW(i,j)}: all points of object i are below all points of object j in an axis normal to the table plane (see Eq. \ref{eq:sparse_rew_BELOW})
    \item \textit{LEFT(i,j)}: all points of object i are bigger than all points of object j in an axis parallel to the x axes of the table plane (see Eq. \ref{eq:sparse_rew_LEFT})
    \item \textit{RIGHT(i,j)}: all points of object i are smaller than all points of object j in an axis parallel to the x axes of the table plane (see Eq. \ref{eq:sparse_rew_RIGHT})
    \item \textit{ABOVECLOSE(i,j)}, \textit{BELOWCLOSE(i,j)}, \textit{LEFTCLOSE(i,j)}, \textit{RIGHTCLOSE(i,j)}: combination of relational reward structures and \textit{CLOSE(i,j)}
    (see Eq. \ref{eq:sparse_rew_ABOVEC}, \ref{eq:sparse_rew_BELOWC},
    \ref{eq:sparse_rew_LEFTC}, \ref{eq:sparse_rew_RIGHTC})
    \item \textit{ABOVECLOSEBOX(i)}: \textit{ABOVECLOSE(i,box object)}
\end{itemize}

We define the auxiliary reward structures, so that we can - in principle - compute all the required information from one or two image planes (two cameras looking at the workspace). 
Replacing the world coordinates referenced above with pixel coordinates.

In the following equations a definition of all rewards is given. Let $d(o_i, o_j)$ be the distance between the center of mass of the two objects, $\max_a(o_i)$ and $\min_a(o_i)$ denote the maximal (or minimal) pixel locations covered by object i in axis $a \in \{x,y,z\}$.

\begin{equation}
r_{C(i,j)}(\bs, \ba) = \begin{cases}
 1 &\text{iff } \ d(o_i, o_j) \leq 10cm \\
 0 &\text{else},
 \end{cases}
 \label{eq:sparse_rew_CLOSE}
\end{equation}

\begin{equation}
r_{A(i,j)}(\bs, \ba) = \begin{cases}
 1 &\text{iff } \ \max_z(o_j) - \min_z(o_i) \leq 0 \\
 0 &\text{else},
 \end{cases}
 \label{eq:sparse_rew_ABOVE}
\end{equation}

\begin{equation}
r_{AC(i,j)}(\bs, \ba) = r_{A(i,j)}(\bs, \ba) * r_{C(i,j)}(\bs, \ba)
 \label{eq:sparse_rew_ABOVEC}
\end{equation}

\begin{equation}
r_{L(i,j)}(\bs, \ba) = \begin{cases}
 1 &\text{iff } \ \max_x(o_j) - \min_x(o_i) \leq 0 \\
 0 &\text{else},
 \end{cases}
 \label{eq:sparse_rew_LEFT}
\end{equation}

\begin{equation}
r_{LC(i,j)}(\bs, \ba) = r_{L(i,j)}(\bs, \ba) * r_{C(i,j)}(\bs, \ba)
 \label{eq:sparse_rew_LEFTC}
\end{equation}

\begin{eqnarray}
r_{B(i,j)}(\bs, \ba) = r_{A(j,i)}(\bs, \ba)
 \label{eq:sparse_rew_BELOW} \\
r_{R(i,j)}(\bs, \ba) = r_{L(j,i)}(\bs, \ba)
 \label{eq:sparse_rew_RIGHT} \\
r_{BC(i,j)}(\bs, \ba) = r_{AC(j,i)}(\bs, \ba)
 \label{eq:sparse_rew_BELOWC} \\
r_{RC(i,j)}(\bs, \ba) = r_{LC(j,i)}(\bs, \ba)
 \label{eq:sparse_rew_RIGHTC} \\
\end{eqnarray}

In addition to these 'object centric' rewards, we define MOVE, TOUCH and NOTOUCH as:

\begin{equation}
r_{MOVE(i)}(\bs, \ba) = \begin{cases}
|v(o_i)| &\text{iff } \ |v(o_i)| \geq 3\frac{mm}{s} \\
 0 &\text{else},
 \end{cases}
 \label{eq:sparse_rew_MOVE}
\end{equation}

\begin{equation}
r_{T(i)}(\bs, \ba) = \begin{cases}
|\sum_{i \in 1,2,3}{f_i}| &\text{iff } \ |\sum_{i \in 1,2,3}{f_i}| \leq 1N \\
 1 &\text{else},
 \end{cases}
 \label{eq:sparse_rew_TOUCH}
\end{equation}

\begin{equation}
r_{NT(i,j)}(\bs, \ba) = \begin{cases}
 1 &\text{iff } \ r_{T(i)}(\bs, \ba) \leq 0.1 \\
 0 &\text{else},
 \end{cases}
 \label{eq:sparse_rew_NOTOUCH}
\end{equation}

Two objects were used in the experiments, yielding a set of 13 general auxiliary rewards that are used in all simulation experiments.

\subsubsection{External Task Rewards}
For the extrinsic or task rewards we use the notion of STACK(i), for a sparse reward signal that describes the property of an object to be stacked.
As a proxy in simulation we use the collision points of different objects in the scene to determine this reward.
where $col(o_i, o_j) = 1$ if object i and j in simulation do have a collision -- 0 otherwise. We can derive a simple sparse reward from these signals as

\begin{equation}
r_{STACK(i)}(\bs, \ba) = \begin{cases}
 1 &\text{iff } \ (1 - col(GROUND, o_i)) \\
  & * (1 - col(ROBOT, o_i)) \\
  & * col(o_j, o_i) = 0\\
 0 &\text{else}.
 \end{cases}
 \label{eq:sparse_rew_STACK}
\end{equation}

For the cleanup experiments we use an additional auxiliary reward for each object, ABOVE\_CLOSE\_BOX (ACB), that accounts for the relation between the object and the box:

\begin{eqnarray}
r_{ACB(i)}(\bs, \ba) = r_{AC(i,BOX)}(\bs, \ba).
\end{eqnarray}

As additional extrinsic reward, we use a sparse INBOX(i) reward signal, that gives a reward of one if the object i is in the box; INBOXALL, that gives a signal of 1 only if all objects are in the box; and a OPENBOX, which yields a sparse reward signal when the lid of the box is lifted higher then a certain threshold,

\begin{equation}
r_{INBOX(i)}(\bs, \ba) = \begin{cases}
 1 &\text{iff } \ o_i \text{ is in box}\\
 0 &\text{else},
 \end{cases}
 \label{eq:sparse_rew_INBOX}
\end{equation}

\begin{equation}
r_{INBOXALL}(\bs, \ba) = \begin{cases}
 1 &\text{iff } \ \text{all objects in box}\\
 0 &\text{else},
 \end{cases}
 \label{eq:sparse_rew_INBOXALL}
\end{equation}

\begin{equation}
r_{OPENBOX}(\bs, \ba) = \begin{cases}
 1 &\text{iff } \ \theta_{lid} \geq 1.5 \\
 0 &\text{else},
 \end{cases}
 \label{eq:sparse_rew_OPENBOX}
\end{equation}

This gives 15 auxiliary reward signals and 4 extrinsic reward signals for the cleanup experiment.

\subsection{Real Robot}

On the real robot we use a slightly altered set of auxiliary rewards to account for the fact that the robot does not possess touch sensors (so TOUCH and NOTOUCH cannot be used), and to reduce the amount of training time needed (a distance based reward for reaching is added for this reason). 
For the pick up experiment we used the following rewards: OPENED, CLOSED, LIFTED(block) and AT(hand,block), defined as:

\begin{itemize}

\item \textit{OPENED, CLOSED}: maximal if the angle of the finger motors, $\theta_{fingers} \in [0.0, ~0.8]$, is close to its minimum respectively maximum value. (see Eq. \ref{eq:real_rew_OPENED} and \ref{eq:real_rew_CLOSED})

\item \textit{LIFTED(i)}: maximal if the lowest point of object i is at a height of 7.5cm above the table, with a linear shaping term below this height. (see Eq. \ref{eq:real_rew_LIFTED})

\item \textit{AT(i, j)}: similar to CLOSE(i,j) in simulation but requiring objects to be closer; maximal if the centers of i and j are within 2cm of each other; additionally uses a non-linear shaping term when further apart. (equivalent to CLOSE$_{1cm}$(i,j) in Eq. \ref{eq:real_rew_REACH})

\end{itemize}

The rewards are defined as followed:

\begin{equation}
r_{OPENED}(\bs, \ba) =
\begin{cases}
  1 &\text{iff} \ \theta_{fingers} \leq 0.1 \\
  0 &\text{else},
\end{cases}
\label{eq:real_rew_OPENED}
\end{equation}

\begin{equation}
r_{CLOSED}(\bs, \ba) =
\begin{cases}
  1 &\text{iff} \ \theta_{fingers} \geq 0.7 \\
  0 &\text{else},
\end{cases}
\label{eq:real_rew_CLOSED}
\end{equation}

\begin{equation}
r_{LIFTED(i)}(\bs, \ba) =
\begin{cases}
  1.5 &\text{iff} \ min_z(i) > 7.5cm \\
  0. &\text{iff} \ min_z(i) < ~0.5cm \\
  \frac{min_z(i)}{7.5} &\text{else},
\end{cases}
\label{eq:real_rew_LIFTED}
\end{equation}

For all other rewards based on the relation between two entities i and j, we use a shaped variant of CLOSE that is parametrized by a desired distance $\epsilon$. Let $d(i, j)$ be the distance between the center $i$ and some target site $j$.

\begin{equation}
r_{CLOSE_x(i,j)}(\bs, \ba) =
\begin{cases}
  1.5 &\text{iff} \ d(i,j) < \epsilon \\
  1 - tanh^2(\frac{d(i,j)}{10}) &\text{else},
\end{cases}
\label{eq:real_rew_REACH}
\end{equation}

In an extended experiment, the agent is trained to bring the object to a specified target position, as well as to hover it above it. For this, we added several more rewards based on a fixed target site.
\begin{itemize}

\item \textit{CLOSE(i, j), AT(i, j)}: maximal if the center of object i is within 10cm respectively 1.5cm of the target j. (equivalent to CLOSE$_{10cm}$(i,j) and CLOSE$_{1.5cm}$(i,j) in Eq. \ref{eq:real_rew_REACH})

\item \textit{ABOVE$\_$CLOSE(i, j), ABOVE$\_$AT(i, j)}: maximal if the center of object i is within 10cm respectively 2cm of a site 6cm above the target j. (equivalent to CLOSE$_{10cm}$(i,j+6cm) and CLOSE$_{2cm}$(i,j+6cm) in Eq. \ref{eq:real_rew_REACH})

\end{itemize}

\section{Additional model details}

For the SAC-X experiments we use a shared network architecture to instantiate the policy for the different intentions. The same basic architecture is also used for the critic Q value function. Formally, $\theta$ and $\phi$ in the main paper thus consist of the parameters of these two neural networks (and gradients for individual intentions wrt. these model parameters are averaged).

\begin{figure}[ht]
\vskip 0.2in
\begin{center}
\centerline{\includegraphics[width=\columnwidth]{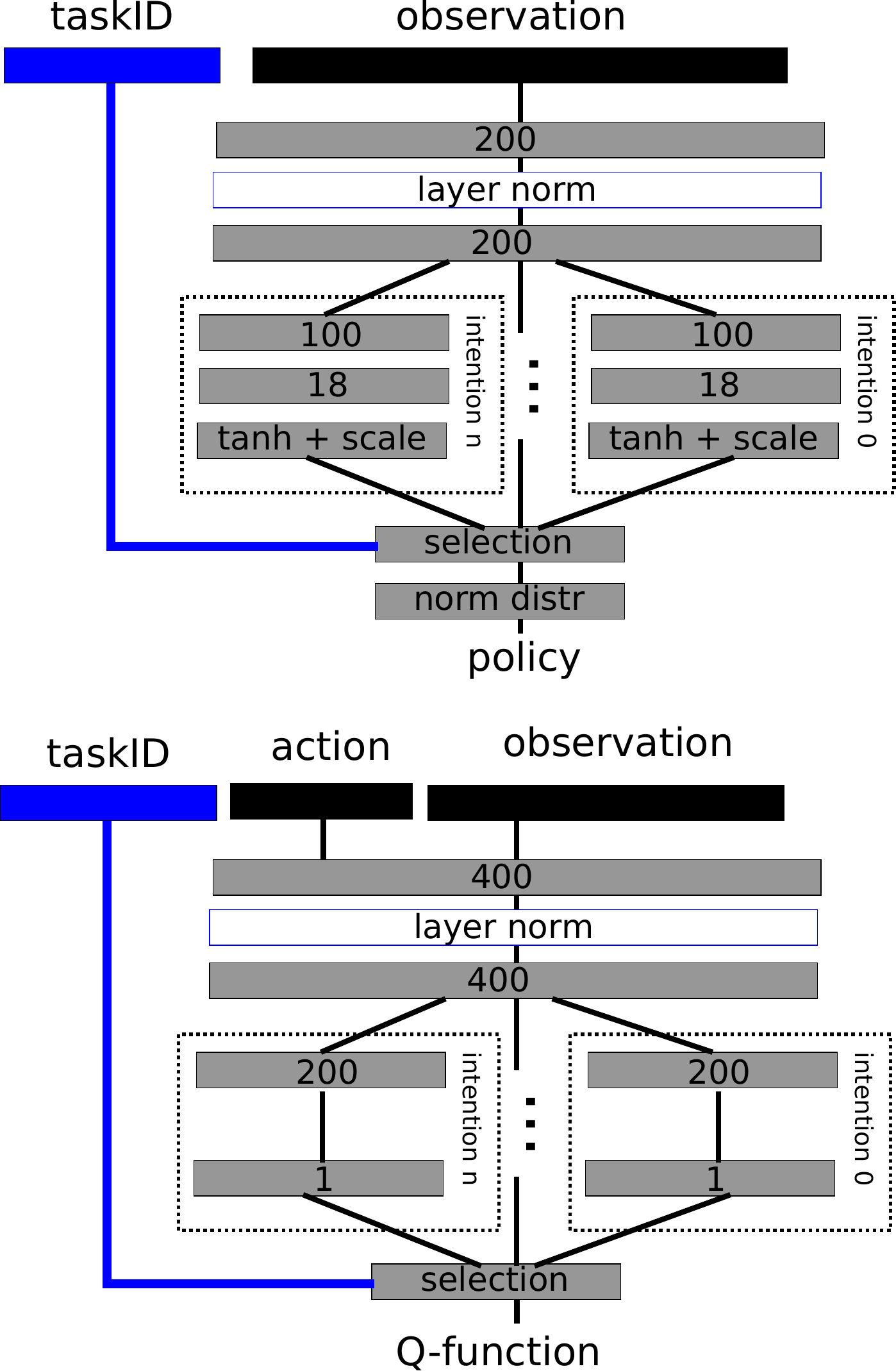}}
\caption{Schematics of the fully connected networks used to parameterize policy distribution and Q-functions for each intention.}
\label{fig:sac_networks}
\end{center}
\vskip -0.2in
\end{figure}

In detail: the stochastic policy consists of a layer of 200 hidden units with ELU units \citep{ClevertUH15}, that is shared across all intentions. After this first layer a LayerNorm \citep{BaLayerNorm16} is placed to normalize activations (we found this to generally be beneficial when switching between different environments that have differently scaled observations). The LayerNorm output is fed to a second shared layer with 200 ELU units. 
The output of this shared stack is routed to blocks of 100 and 18 ELU units followed by a final tanh activation. This output determines the parameters for a normal distributed policy with 9 outputs (whose variance we allow to vary between 0.3 and 1 by transforming the corresponding tanh output accordingly).
For the critic we use the same architecture, but with 400 units per layer in the shared part and a 200-1 head for each intention. Figure \ref{fig:sac_networks} shows a depiction of this model architecture.
For the pixel based experiments a CNN stack consisting of two convolutional layers (16 feature maps each, with a kernel size of 3 and stride 2) processes two, stacked, input images of 48 x 48 pixels. The output of this stack is fed to a 200 dimensional linear layer (again with ELU activations) and concatenated to the output of the first layer in the above described architecture (which now only processes proprioceptive information).

The intentions are 1 hot encoded and select which head of the network is active for the policy and the value function.
Other network structures (such as feeding the selected intention into the network directly) worked in general, but the gating architecture described here gave the best results -- with respect to final task performance -- in preliminary experiments.

Training of both policy and Q-functions was performed via ADAM \citep{Kingma2015Adam} using a learning rate of $2 \cdot 10^{-4}$ (and default parameters otherwise). See also the next section for details on the algorithm.

\subsection{Stochastic Value Gradient for Learned Intentions}
The following presents a detailed derivation of the stochastic value gradient -- Equation (9) in the main paper -- for learning the individual intention policies.
Without loss of generality, we assume Gaussian policies for all intentions (as used in all our experiments). We can then first reparaemeterize the sampling process for policy $\ba_t \sim \pi_\btheta(\ba | \bs_t, \cT)$ as
$g_\theta(\bs_t, \epsilon_\ba)$, where $\epsilon_\ba$ is a random variable drawn from an appropriately chosen base distribution. That is, for a Gaussian policy we can use a normal distribution~\citep{Kingma2013auto,rezende14} $\epsilon_\ba \sim \mathcal{N}(\mathbf{0}, \mathbf{I})$, with $I$ denoting the identity matrix. 
More precisely, let $\sim \pi_\btheta(\ba | \bs_t, \cT) = \mathcal{N}(\mu_\theta(\bs_t), \sigma^2_\theta(\bs_t))$, then $g_\theta(\bs_t, \epsilon_a) = \mu_\theta(\bs_t) + \sigma_\theta(\bs_t) * \epsilon_\ba$.
With this definition in place we can re-write the gradient as
\begin{equation}
\begin{aligned}
\nabla_\btheta &\cL(\btheta) \\ \approx &\sum_{\substack{\cT \in \sT \\ \tau \sim B}} \mathop{\nabla_{\btheta} \bE}_{\substack{\pi_\btheta(\cdot | \bs_t, \cT) \\ \bs_t \in \tau}} \Big\lbrack \hat{Q}^{\pi}_\cT(\bs_t, \ba; \phi) + \alpha \log \pi_\btheta(\ba | \bs_t, \cT) \Big\rbrack, \\
= &\sum_{\substack{\cT \in \sT \\ \tau \sim B}} \mathop{\bE}_{\substack{\epsilon_\ba \sim \mathcal{N}(\mathbf{0}, \mathbf{I})\\ \bs_t \in \tau}} \Big\lbrack \nabla_{g} \hat{Q}^{\pi}_\cT(\bs_t, g_\theta(\bs_t, \epsilon_\ba); \phi) \nabla_{\btheta} g_\theta(\bs_t, \epsilon_\ba) \\
&+ \alpha \nabla_g \log \pi_\btheta(g(\bs_t, \epsilon_\ba) | \bs_t, \cT) \nabla_{\btheta} g_\theta(\bs_t, \epsilon_\ba) \Big\rbrack.
\end{aligned}
\end{equation}

\section{SAC-Q algorithm}
To allow for fast experimentation we implement our algorithm in a distributed manner, similar to recent distributed off-policy implementations from the literature \citep{Gu17,Horgan18}. In particular, we perform asynchronous learning and data acquisition in the following way: 
Except for the real world experiment, in which only a single robot -- one actor connected to $10$ learners -- is used, we launch $36$ actor processes that gather experience. These actors are connected to $36$ learners (we used a simple 1-to-1 mapping) and send experience over at the end of each episode. To allow for fast learning of the scheduling choices each actor also performs Monte Carlo estimation of the Scheduling rollouts (i.e. it keeps its own up-to-date scheduler). The complete procedure executed by each actor is given in Algorithm \ref{alg:actor}.

The learners then aggregate all collected experience inside a replay buffer and calculate gradients for the policy and Q-function networks, as described in Algorithm \ref{alg:learner}. 

Each learner then finally sends gradients to a central parameter server, that collects $G = 36$ gradients, updates the parameters and makes them available for both learners and actors; see the algorithm listing in Algorithm \ref{alg:chief}.

Note that this setup also makes experimentation on a real robot easy, as learning and acting (the part of the procedure that needs to be executed on the real robot) are cleanly separated.

\begin{algorithm}
\caption{SAC-Q (parameter server)}\label{alg:chief}
\begin{algorithmic}
\STATE \textbf{Input:} $G$ number of gradients to average
\STATE Initialize parameters $\theta, \phi$
\WHILE{True}
\STATE initialize N = 0
\STATE initialize gradient storage $d_\theta = \{ \}, d_\phi = \{ \}$
\WHILE{$N < G$}
\STATE receive next gradients from learner $i$
\STATE $d^\cT_\phi = d^\cT_\phi \cup \{\delta\phi^i \}$
\STATE $d_\theta = d_\theta \cup \{\delta\theta^i\}$
\ENDWHILE
\STATE update parameters with averages from gradient store:
\STATE $\phi =$ ADAM\_update($\phi$, $\frac{1}{|d_\phi|} \sum_{\delta_\phi \in d_\phi} \delta_\phi$) 
\STATE $\theta =$ ADAM\_update($\theta$, $\frac{1}{|d_\theta|} \sum_{\delta_\theta \in d_\theta} \delta_\theta$) 
\STATE send new parameters to workers
\ENDWHILE
\end{algorithmic}
\end{algorithm}

\begin{algorithm}
\caption{SAC-Q (learner)}\label{alg:learner}
\begin{algorithmic}
\STATE \textbf{Input:} $N_\text{learn}$ number of learning iterations, $\alpha$ entropy regularization parameter
Fetch initial parameters $\theta, \phi$
\WHILE{$N < N_\text{learn}$}
\STATE update replay buffer $B$ with received trajectories
\FOR{k=0,1000}
\STATE sample a trajectory $\tau$ from $B$ 
\STATE // compute gradients for policy and Q
\STATE $\delta_\phi = \frac{1}{|\sT|} \sum_{\cT \in \sT} \nabla_\phi L(\phi) $
\STATE $\delta_\theta = \nabla_\theta L(\theta)$
\STATE send $(\delta_\theta, \delta_\phi)$ to parameter server
\STATE wait for parameter updates
\STATE fetch new parameters $\phi$, $\theta$
\ENDFOR
\STATE // update target networks
\STATE $\phi' = \phi$, $\theta' = \theta$
\STATE $N = N + 1$
\ENDWHILE
\end{algorithmic}
\end{algorithm}

\begin{algorithm}
\caption{SAC-Q (actor)}\label{alg:actor}
\begin{algorithmic}
\STATE \textbf{Input:} $N_\text{trajectories}$ number of total trajectories requested, $T$ steps per episode, $\xi$ scheduler period
\STATE // Initialize Q-table
\STATE $ \forall \cT_h, \cT_{0:h-1} : Q(\cT_{0:h-1}, \cT_h) = 0$, $M_{\cT_h} = 0$
\WHILE{$N < N_\text{trajectories}$}
\STATE fetch parameters $\theta$
\STATE // collect new trajectory from environment
\STATE $\tau = \lbrace \rbrace, h = 0$
\FOR{t=0,T}
\IF{$t \pmod{\xi} \equiv 0$}
\STATE $\cT_h \sim P_\cS(\cT | \cT_{0:h-1} )$
\STATE $h = h + 1$
\ENDIF
\STATE $a_t \sim \pi_\theta(\ba_0 | \bs_t, \cT_h)$
\STATE // execute action and collect all rewards
\STATE $\bar{r} = \lbrack r_{\cA_1}(\bs_t, \ba_t), \dots, r_{|\sA|}(\bs_t, \ba_t), r_{\cM}(\bs_t, \ba_t) \rbrack$
\STATE $\tau \leftarrow \tau \cup \lbrace (\bs_t, \ba_t, \bar{r}, \pi_\theta(\ba_0 | \bs_t, \cT_h)) \rbrace$
\ENDFOR
\STATE send $\tau$ and schedule decisions $\cT_{0:H}$ to learner 
\STATE // update Monte Carlo Q for scheduler$P_S$ 
\FOR{h=0:H}
\STATE $M_{\cT_h} = M_{\cT_h} + 1$
\STATE $Q(\cT_{0:h-1}, \cT_h) \pluseq  \frac{R^\tau_\cM(\cT_{h:H}) - Q(\cT_{0:h-1}, \cT_h)}{M}$
\ENDFOR
\STATE $N = N + 1$
\ENDWHILE
\end{algorithmic}
\end{algorithm}

\newpage

\section{Additional Experiment Results}
\subsubsection{A detailed look at the SAC-Q learning process}

\begin{figure}[h!]
\vskip 0.2in
\begin{center}
\centerline{\includegraphics[width=\columnwidth]{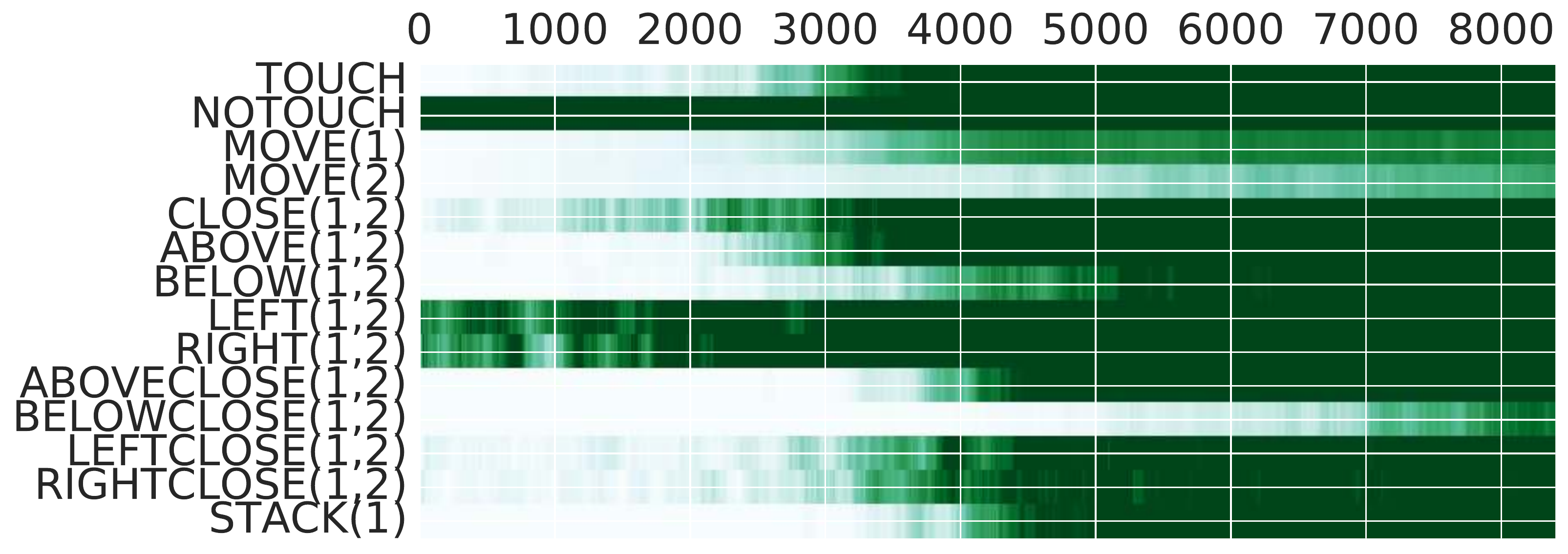}}
\centerline{\includegraphics[width=\columnwidth]{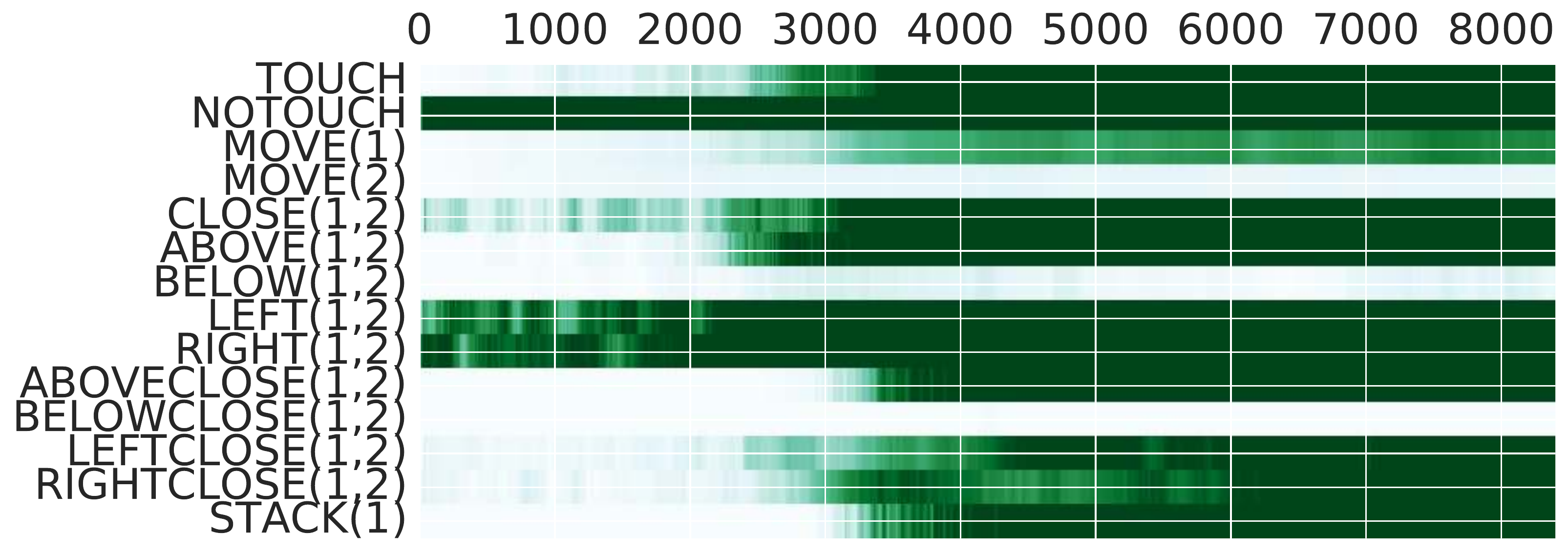}}
\caption{Comparison of full auxiliary and extrinsic set of intentions learned of SAC-U (top) and SAC-Q (bottom) over the training process. The x axis is episodes per actor and the color intensity encodes the obtained reward for each depicted intention.}
\label{fig:stack_1-sac-auxplot-full}
\end{center}
\vskip -0.2in
\end{figure}

In Figure \ref{fig:stack_1-sac-auxplot-full} we show the reward statistics over the full set of auxiliary and extrinsic tasks for both SAC-U (left) and SAC-Q (right) when learning the stacking task.
While our main goal is to learn the extrinsic stacking task, we can observe that the SAC-X agents are able to learn all auxiliary intentions in parallel.
In this example we use a set of 13 auxiliary intentions which are defined on the state of the robot and the two blocks in the scene as in Section \ref{sect:aux_rewards}. These are \textit{TOUCH}, \textit{NOTOUCH}, \textit{MOVE(1)}, \textit{MOVE(2)}, \textit{CLOSE(1,2)}, \textit{ABOVE(1,2)}, \textit{BELOW(1,2)}, \textit{LEFT(1,2)}, \textit{RIGHT(1,2)}, \textit{ABOVECLOSE(1,2)}, \textit{BELOWCLOSE(1,2)}, \textit{LEFTCLOSE(1,2)}, \textit{RIGHTCLOSE(1,2)}.
In addition we have the extrinsic reward, which is defined as \textit{STACK(1)} in this case.
SAC-U (shown in the top part of the figure) will execute all intentions in a uniform order. Some of the intention goals (such as for \textit{NOTOUCH}, \textit{WEST}, \textit{EAST}) can be valid starting states of an episode and will see their reward signals very early in the learning process.
Other reward signals, such as \textit{MOVE} and \textit{TOUCH}, are more difficult to learn and will lead to rich interaction with the environment which are, in turn, a requirement for learning even more difficult intentions.
In this example, after \textit{NORTH} and \textit{NORTHCLOSE} are learned, \textit{PILE(1)} can be learned reliably as well.

\begin{figure}[h!]
\vskip 0.2in
\begin{center}
\centerline{\includegraphics[width=\columnwidth]{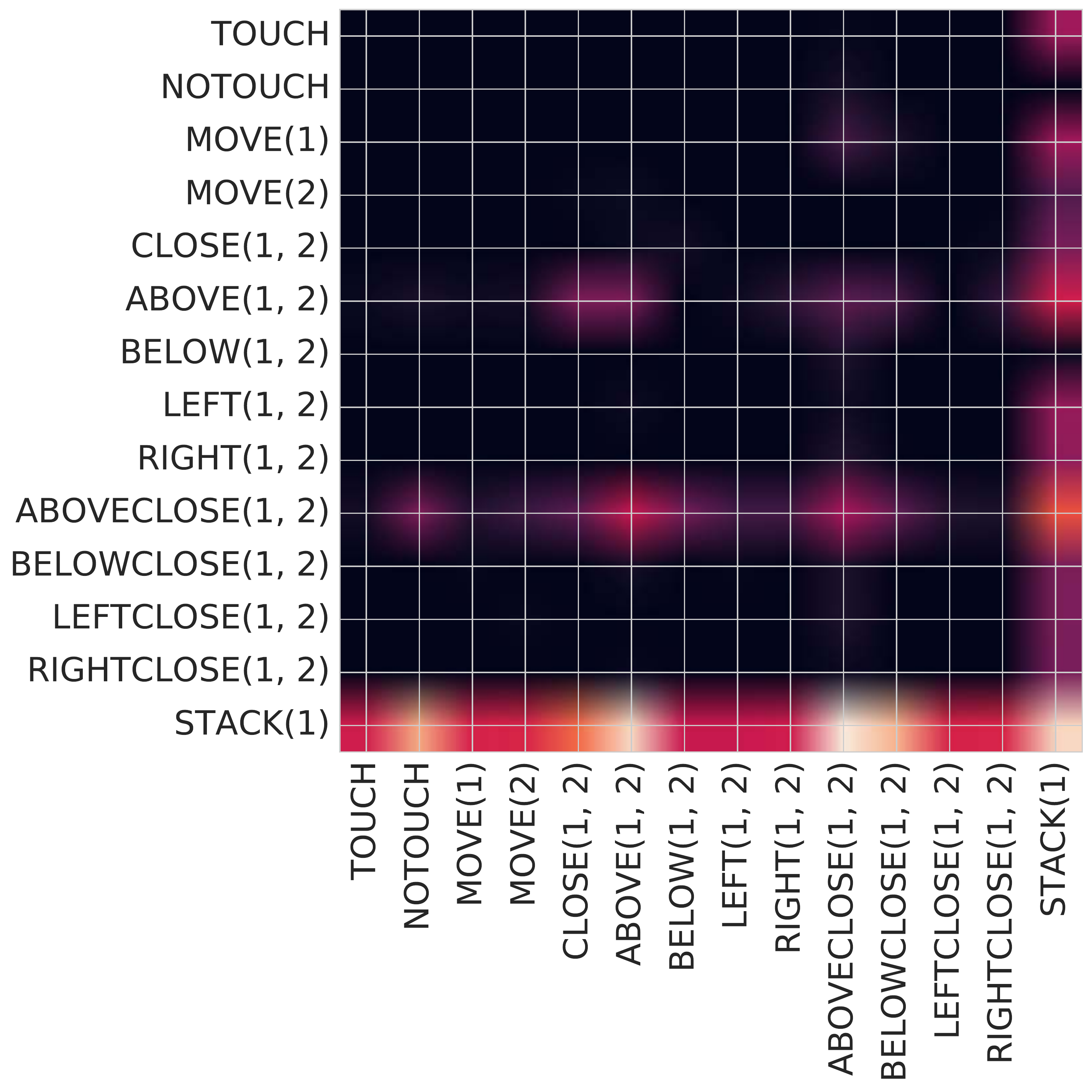}}
\caption{SAC-Q learned Q value distribution for the scheduler. We plot the Q-values after training for pairs of executed intentions. That is, the Q value after first executing the intention denoted by the row names and then executing the intention denoted by the column name. Lighter colors here indicate a higher extrinsic stacking reward.}
\label{fig:stack_1-sac-q-Qplot}
\end{center}
\vskip -0.2in
\end{figure}

The SAC-Q agent in contrast tries to select only auxiliary tasks that will help to collect reward signals for the extrinsic intentions.
In the bottom plot in Figure \ref{fig:stack_1-sac-auxplot-full}, we can see that by ignoring the auxiliaries \textit{MOVE(2)}, \textit{SOUTH} and \textit{SOUTHCLOSE}, SAC-Q manages to learn the extrinsic task faster. 
The learned distribution of Q values at the end of training can also be seen in Figure \ref{fig:stack_1-sac-q-Qplot} (plotted for pairs of executed intentions). We can observe that executing the sequence (\textit{STACK(1)}, \textit{STACK(1)}), gives the highest value, as expected. But SAC-Q also found other sequences of intentions that will help to collect reward signals for \textit{STACK(1)}. 

\begin{figure}[htbp]
\begin{center}
\centerline{\includegraphics[width=\columnwidth]{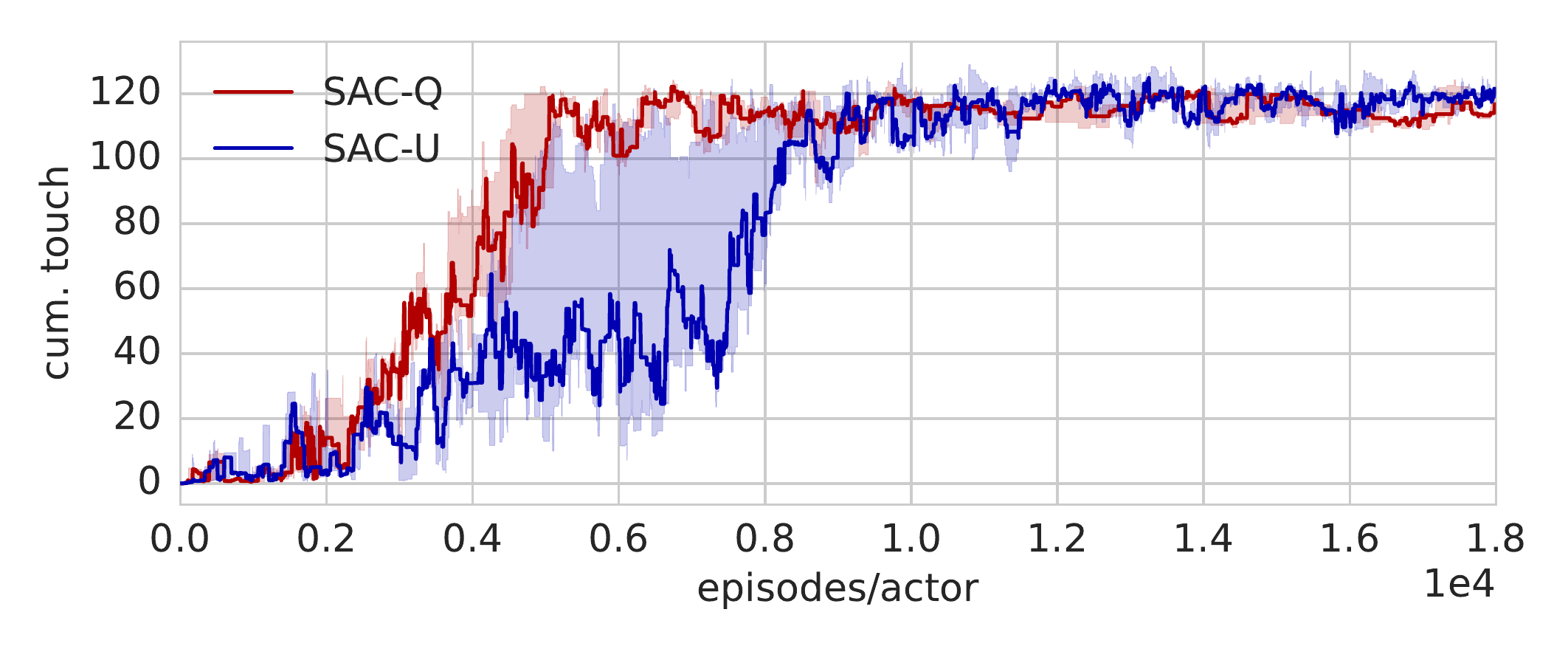}}
\centerline{\includegraphics[width=\columnwidth]{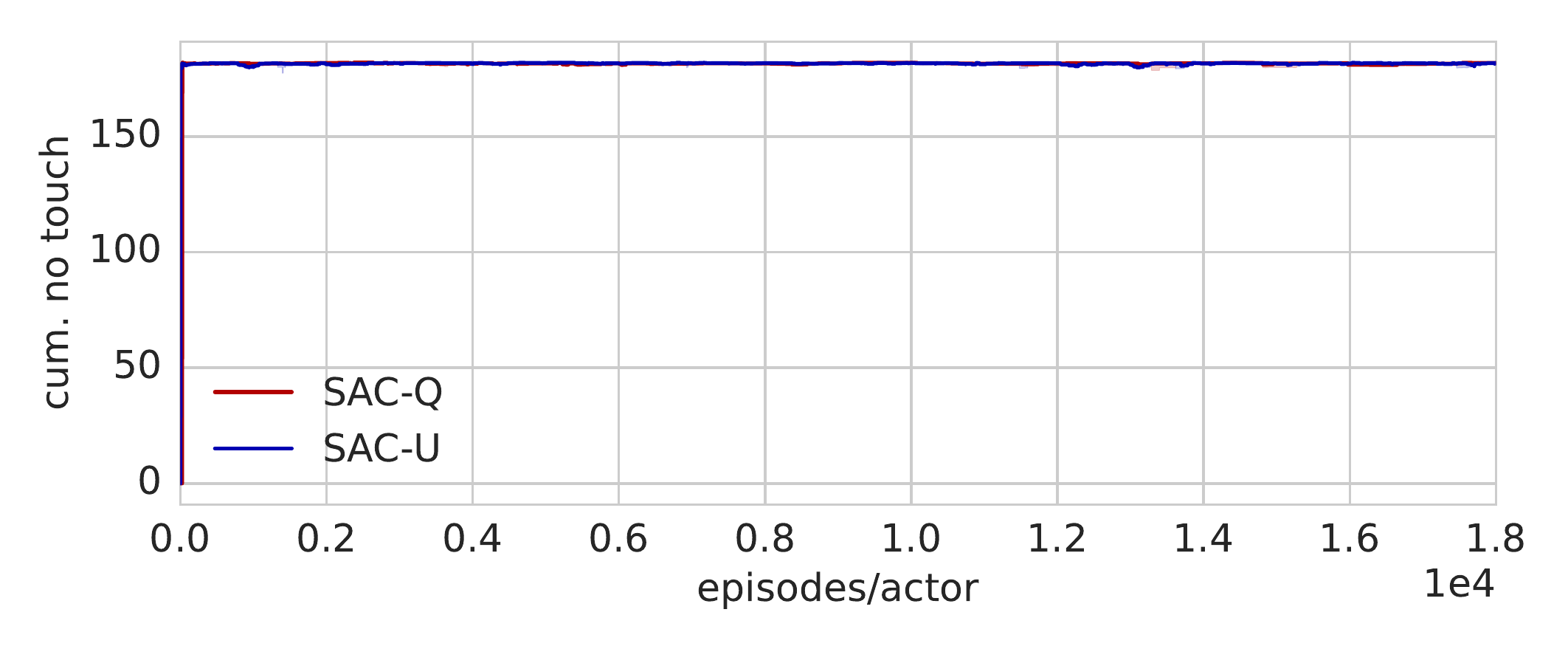}}
\centerline{\includegraphics[width=\columnwidth]{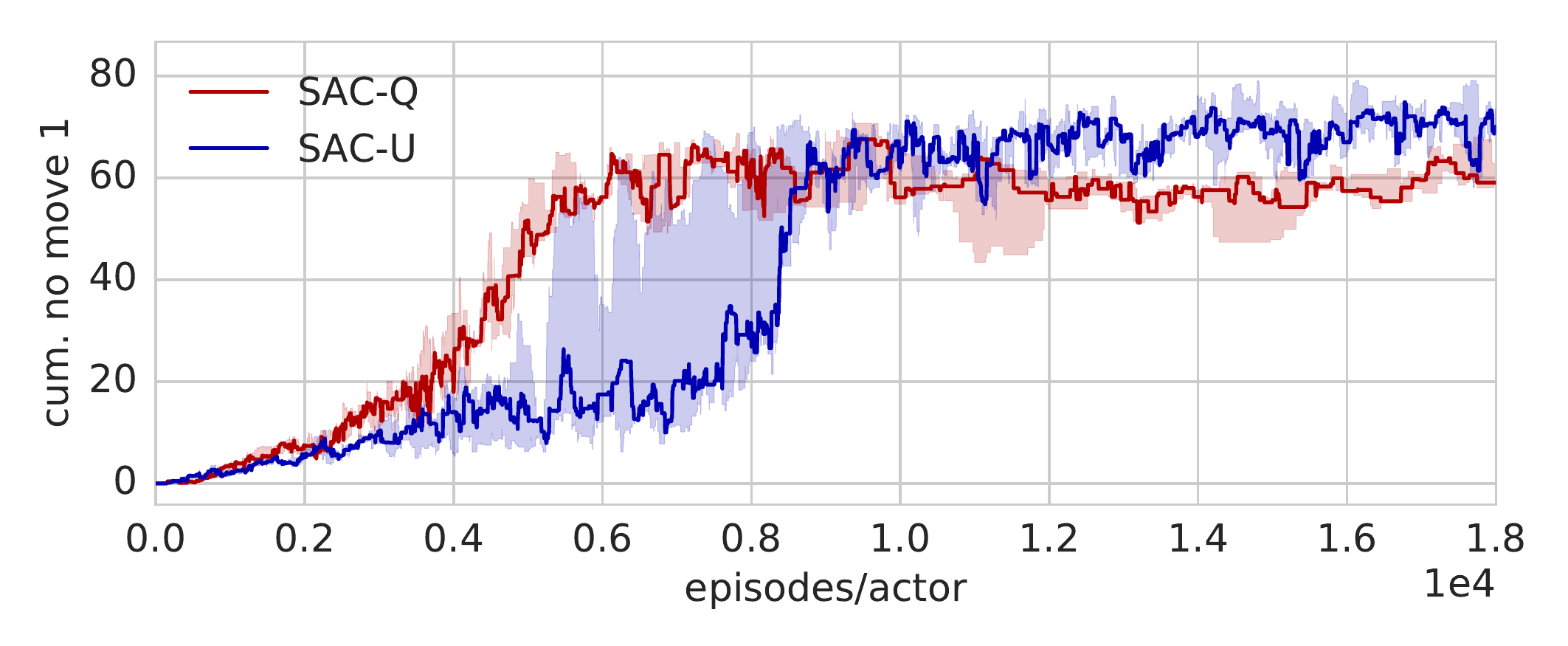}}
\centerline{\includegraphics[width=\columnwidth]{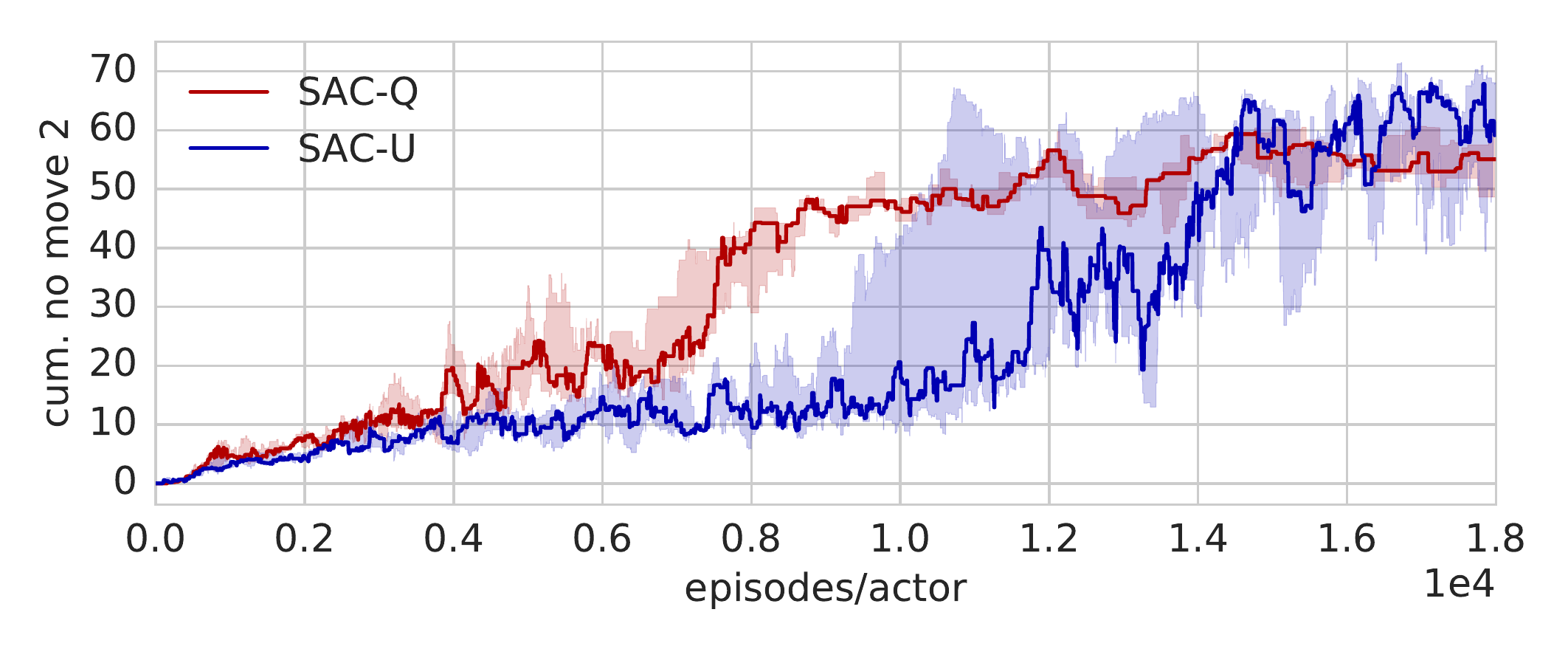}}
\centerline{\includegraphics[width=\columnwidth]{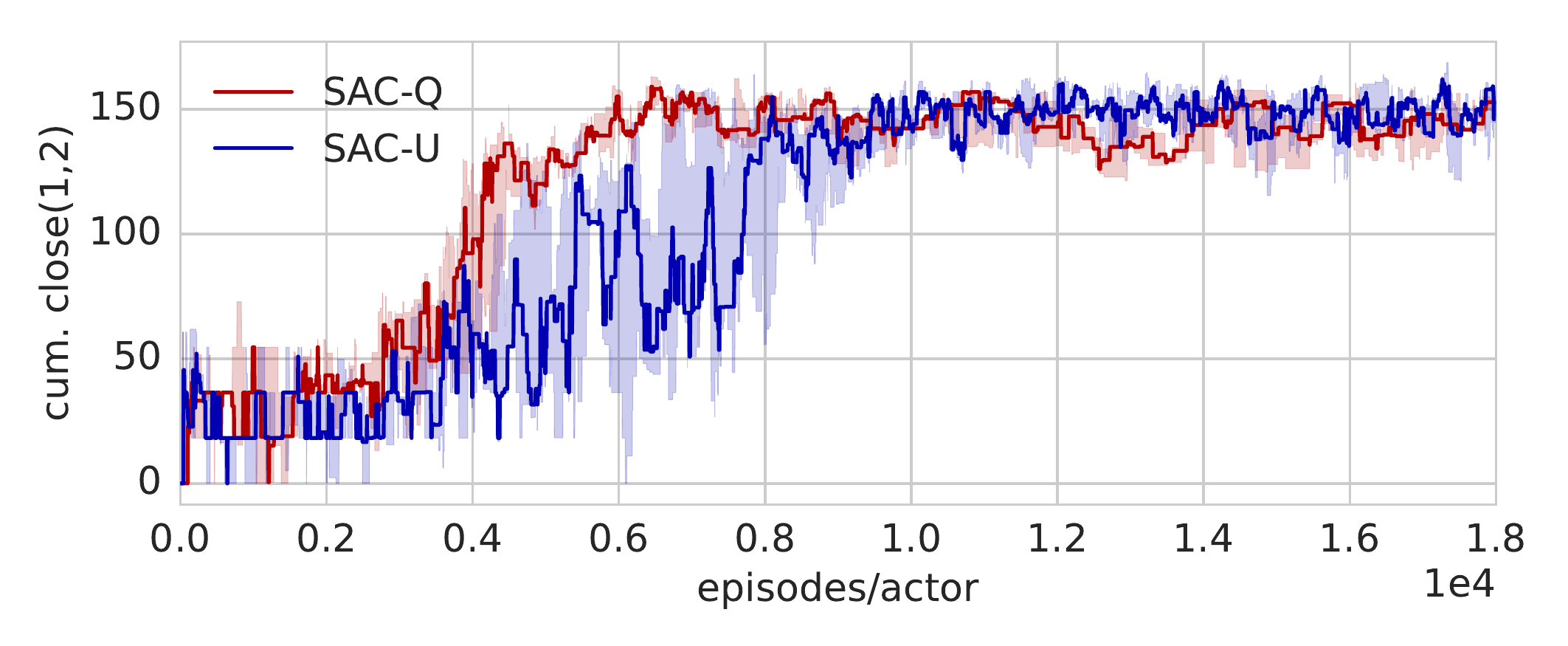}}
\centerline{\includegraphics[width=\columnwidth]{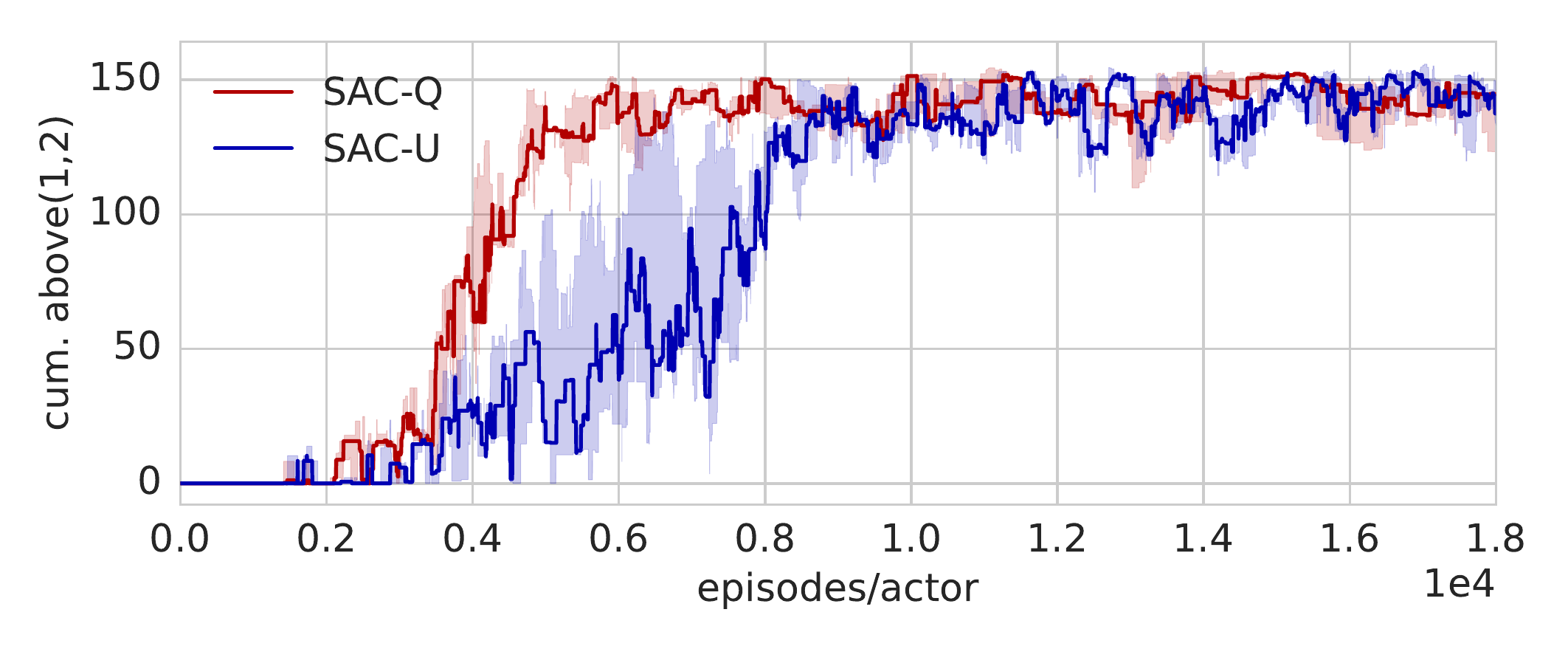}}
\caption{Cleanup experiment, SAC-Q learns all six extrinsic tasks reliably. In addition it reliably learns also to solve the 15 auxiliary tasks in parallel. Part 1: auxiliaries 1-6.}
\label{cleanup-sac-all-1}
\end{center}
\vskip -0.2in
\end{figure}

\begin{figure}[htbp]
\begin{center}
\centerline{\includegraphics[width=\columnwidth]{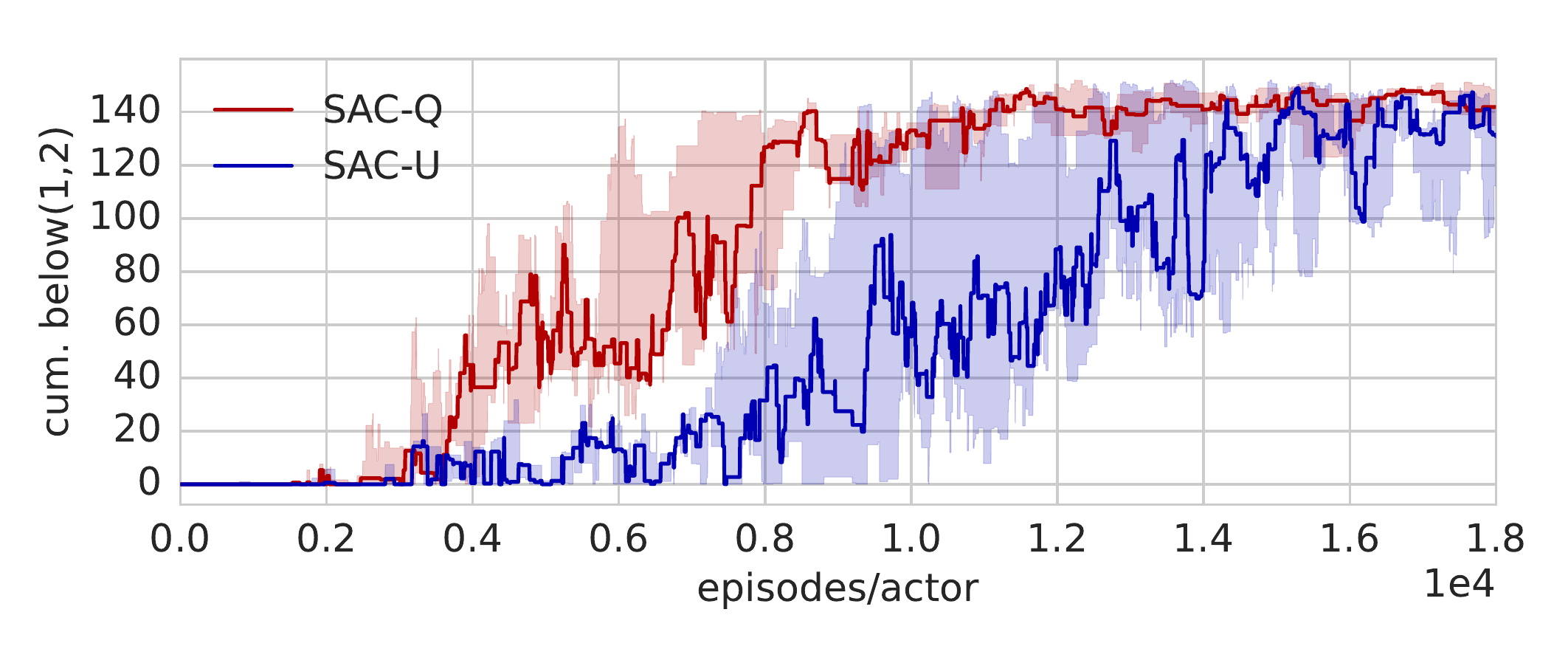}}
\centerline{\includegraphics[width=\columnwidth]{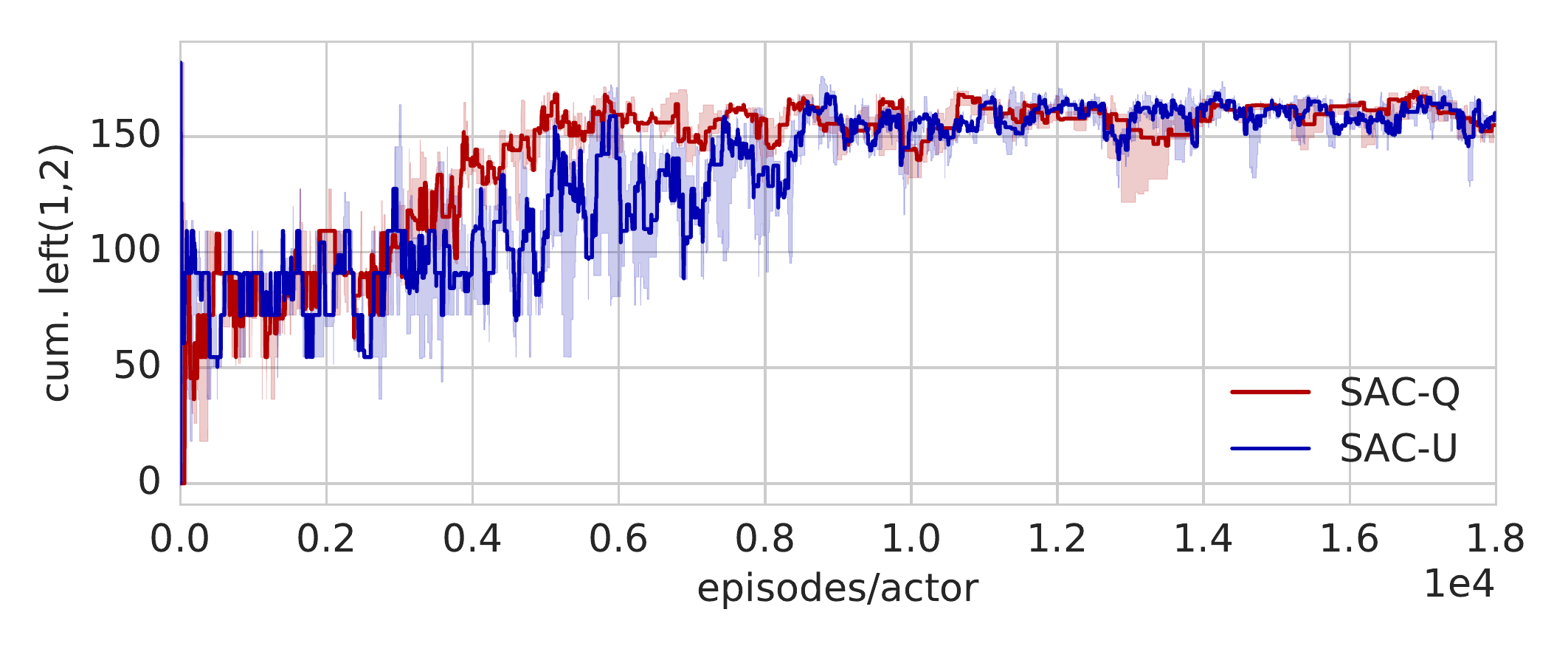}}
\centerline{\includegraphics[width=\columnwidth]{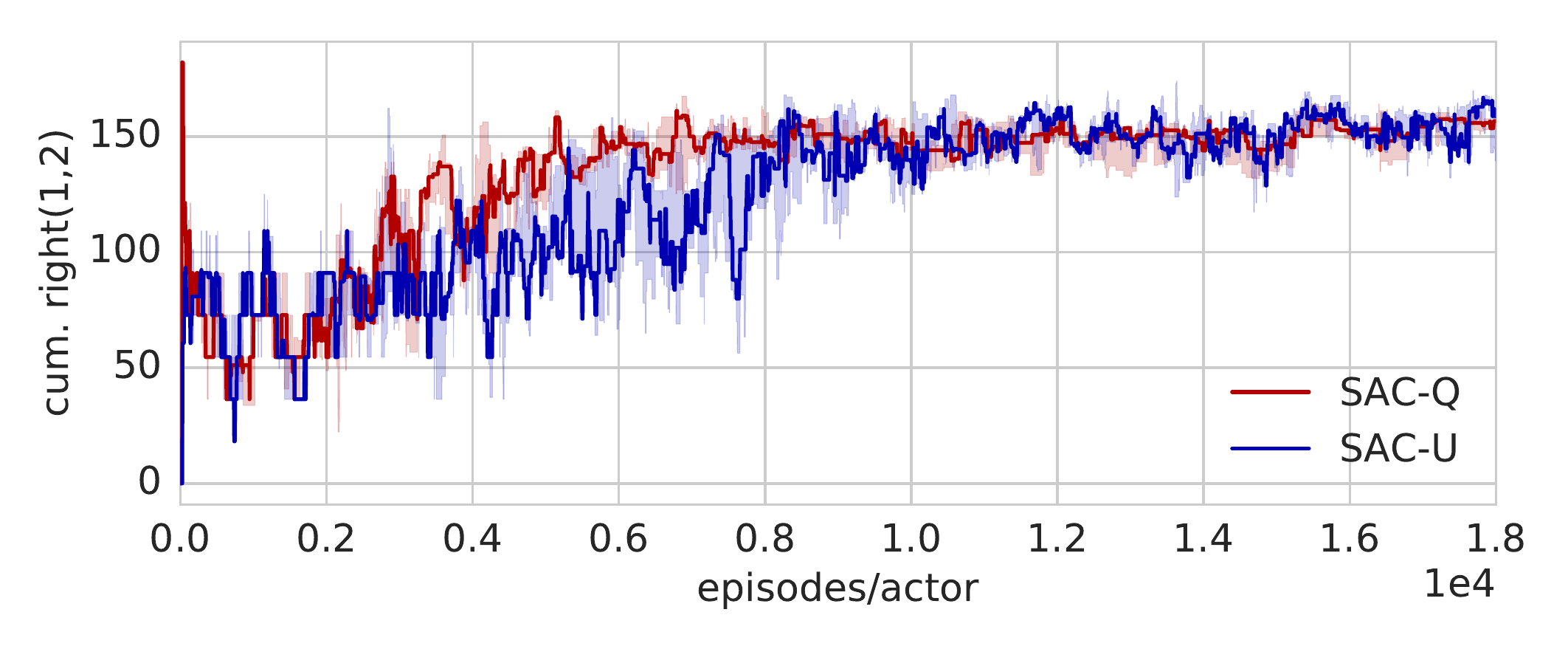}}
\centerline{\includegraphics[width=\columnwidth]{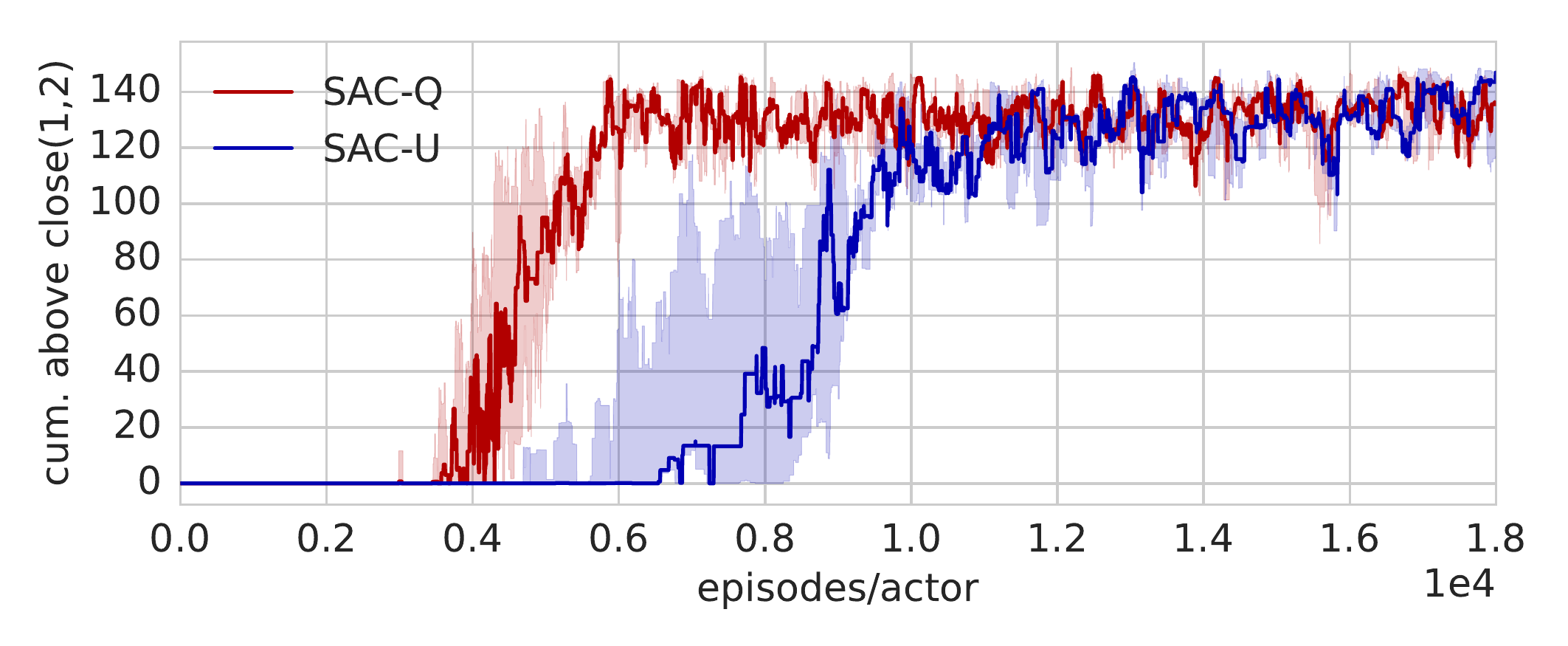}}
\centerline{\includegraphics[width=\columnwidth]{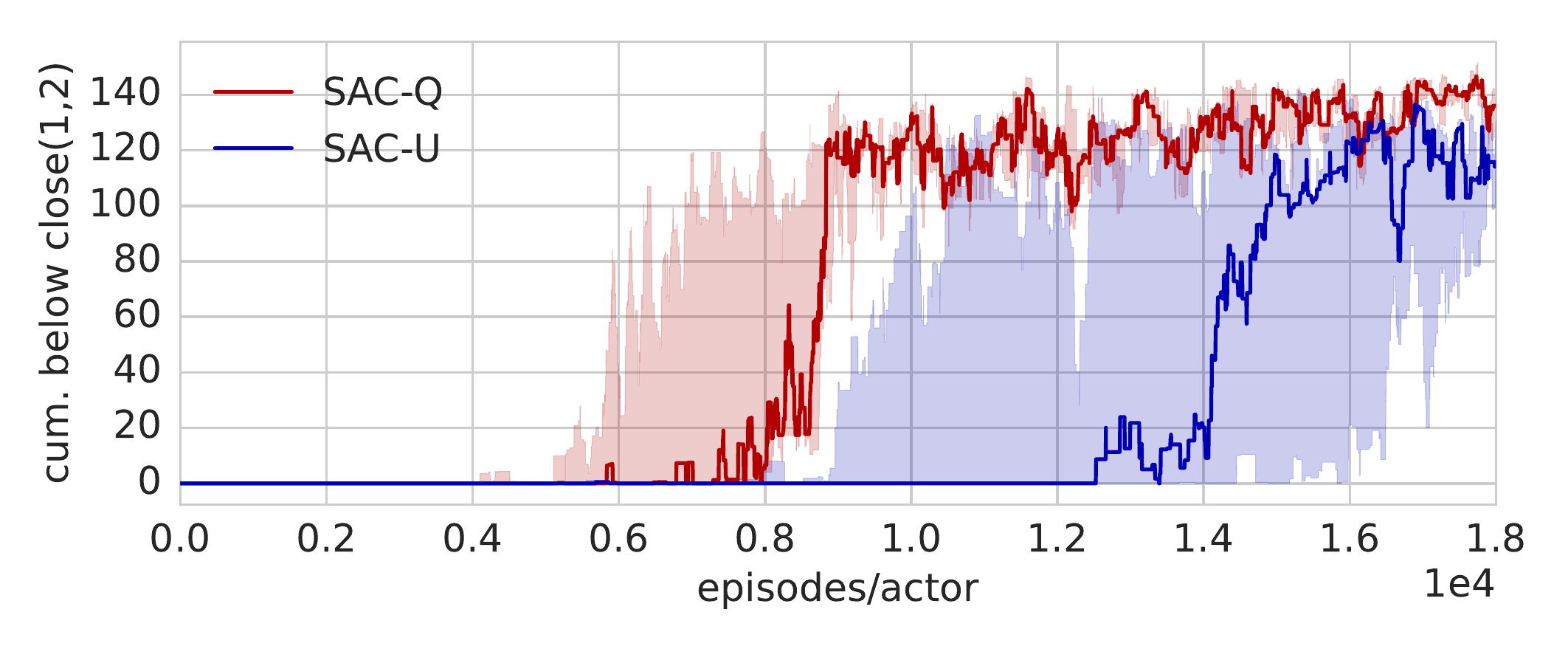}}
\centerline{\includegraphics[width=\columnwidth]{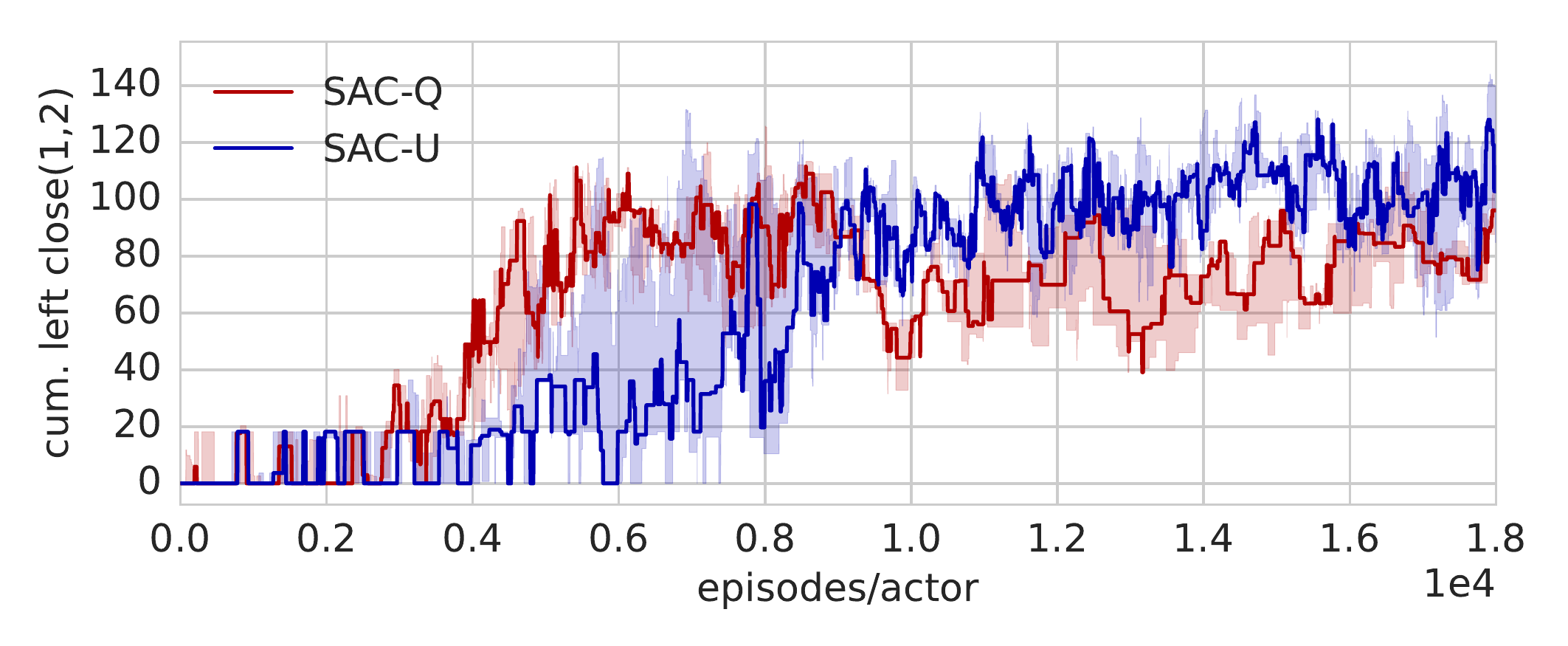}}
\caption{Cleanup experiment, SAC-Q learns all six extrinsic tasks reliably. In addition it reliably learns also to solve the 15 auxiliary tasks in parallel. Part 2: auxiliaries 7-12.}
\label{cleanup-sac-all-2}
\end{center}
\vskip -0.2in
\end{figure}

\begin{figure}[htbp]
\begin{center}
\centerline{\includegraphics[width=\columnwidth]{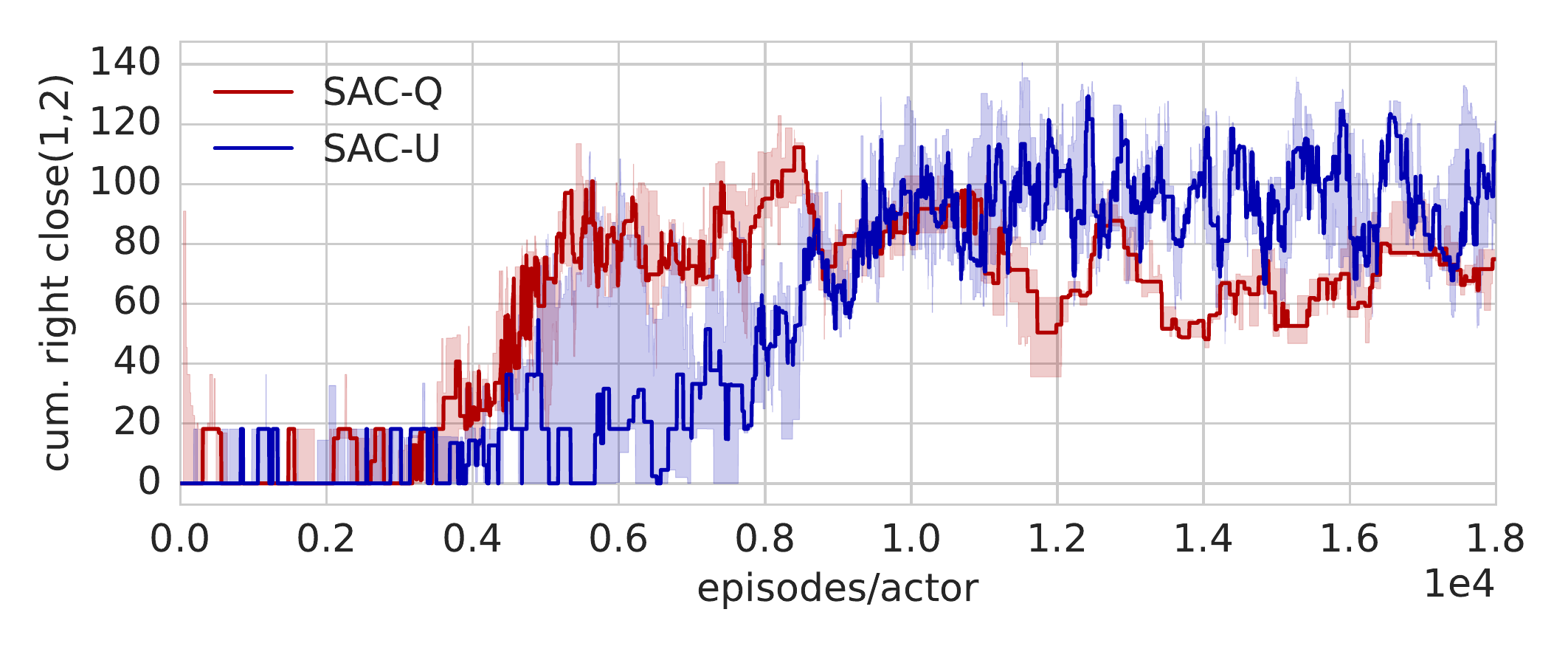}}
\centerline{\includegraphics[width=\columnwidth]{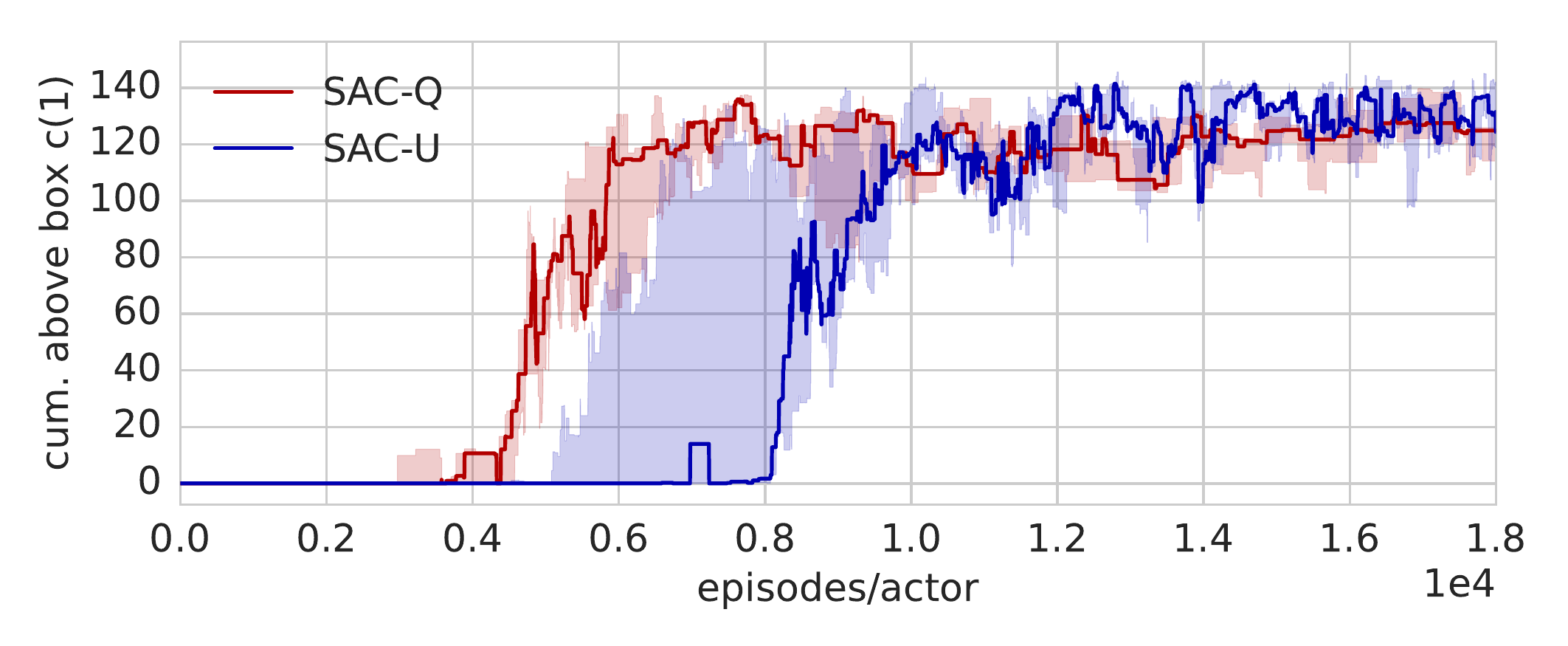}}
\centerline{\includegraphics[width=\columnwidth]{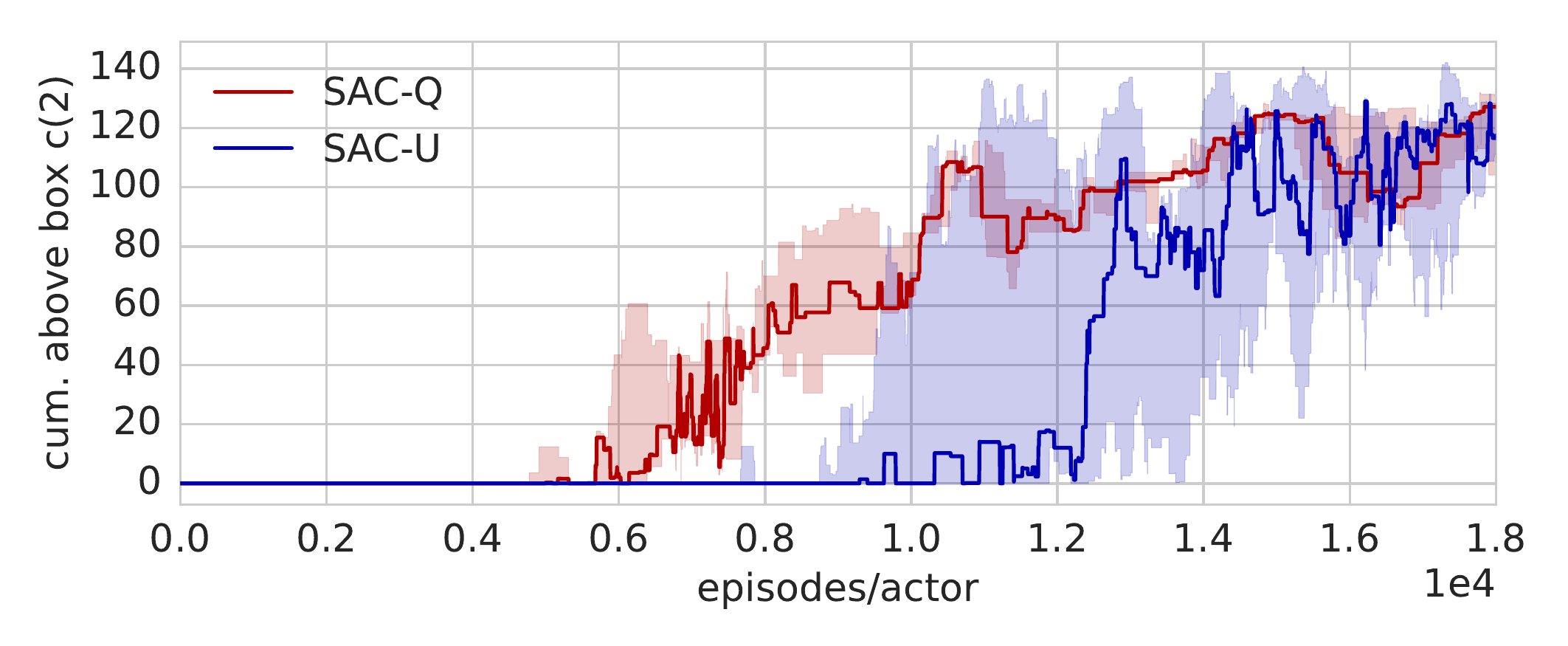}}
\caption{Cleanup experiment, SAC-Q learns all six extrinsic tasks reliably. In addition it reliably learns also to solve the 15 auxiliary tasks in parallel. Part 3: auxiliaries 13-15.}
\label{cleanup-sac-all-3}
\end{center}
\vskip -0.2in
\end{figure}

\begin{figure}[htbp]
\begin{center}
\centerline{\includegraphics[width=\columnwidth]{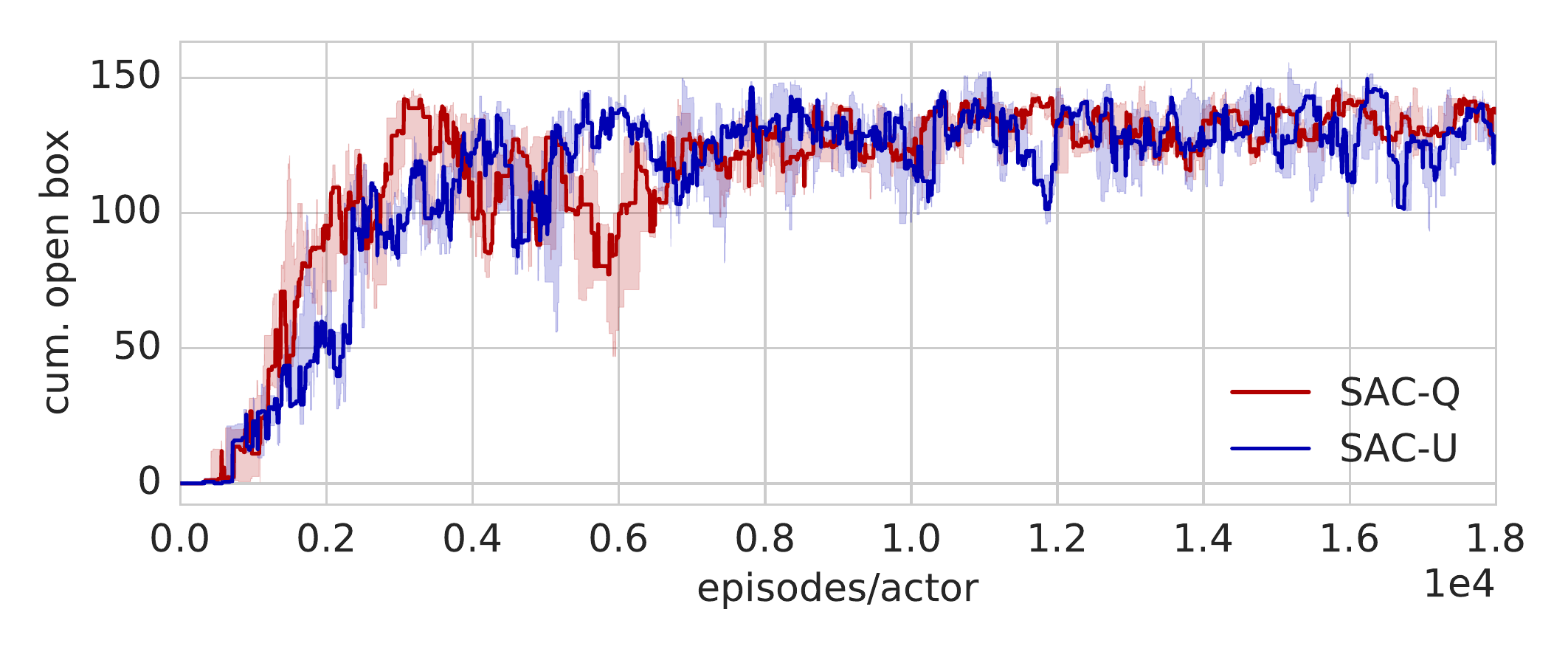}}
\centerline{\includegraphics[width=\columnwidth]{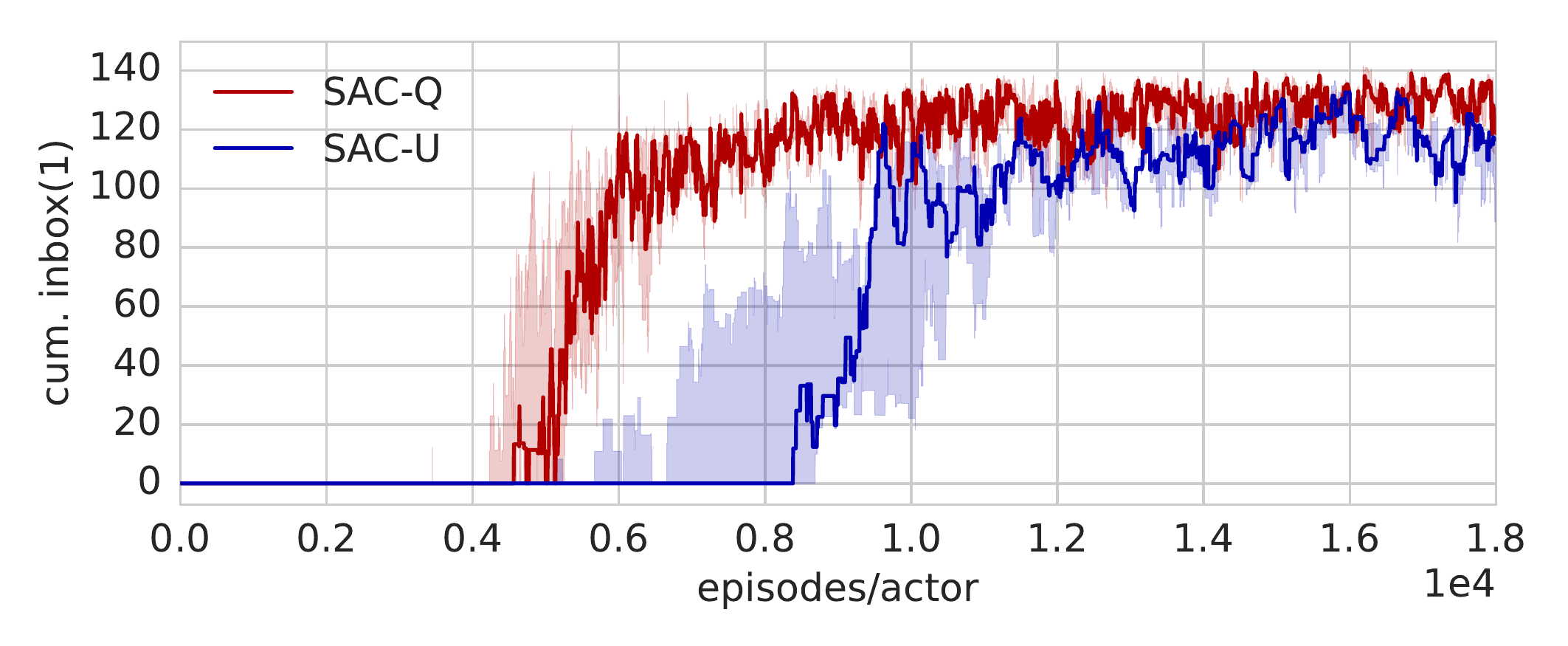}}
\centerline{\includegraphics[width=\columnwidth]{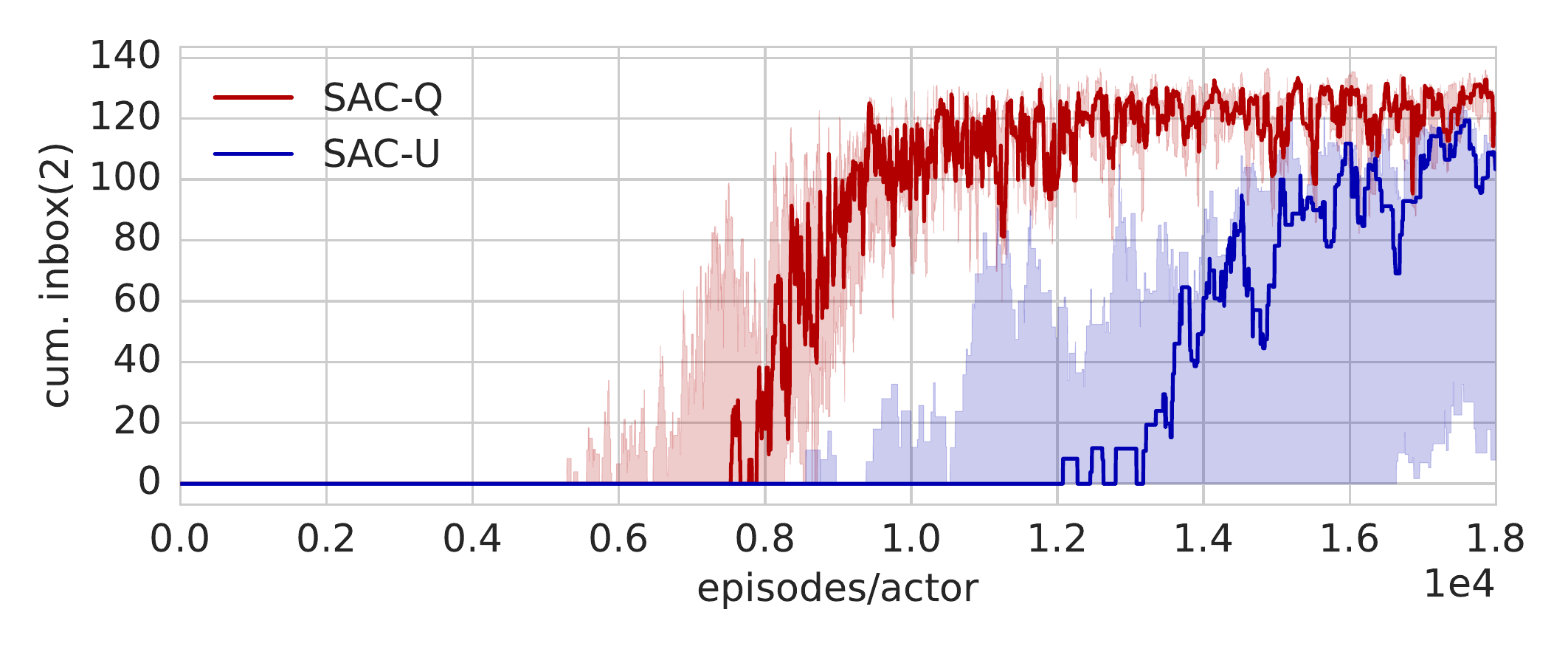}}
\centerline{\includegraphics[width=\columnwidth]{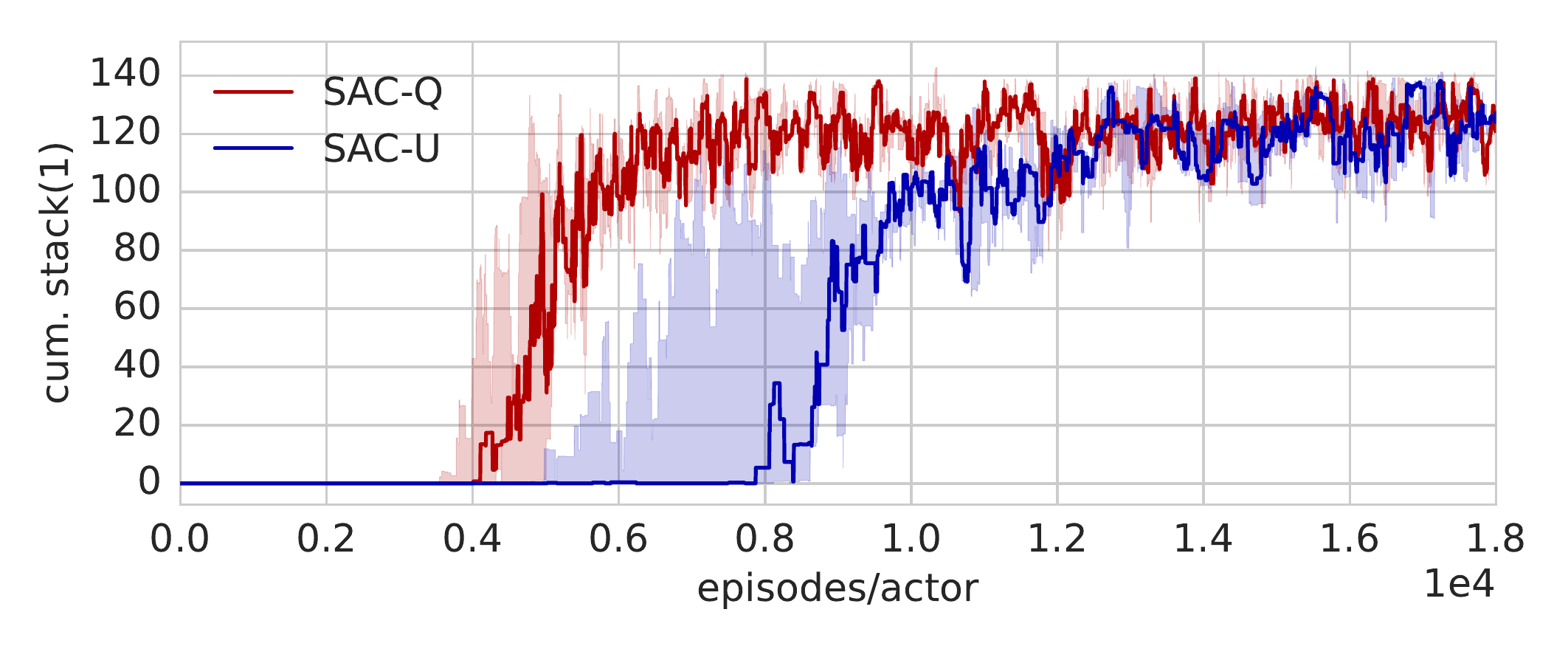}}
\centerline{\includegraphics[width=\columnwidth]{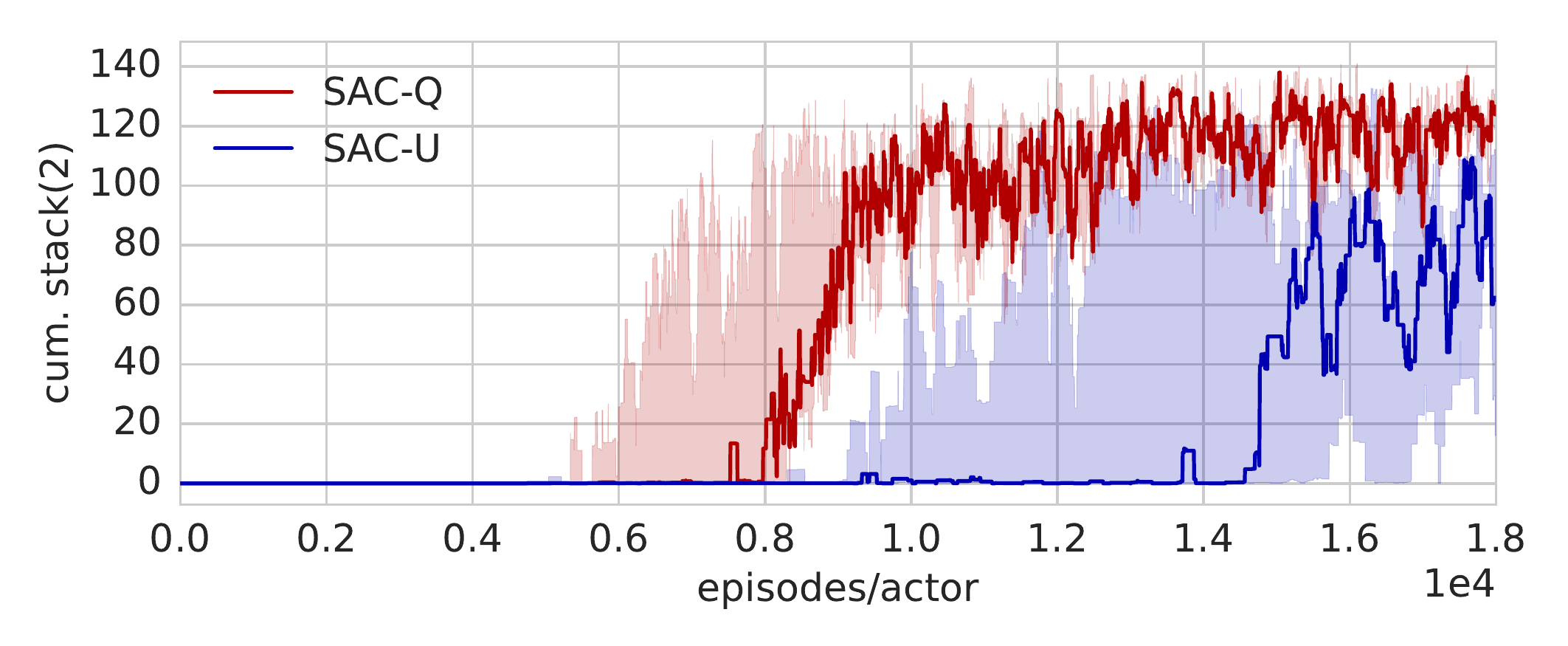}}
\centerline{\includegraphics[width=\columnwidth]{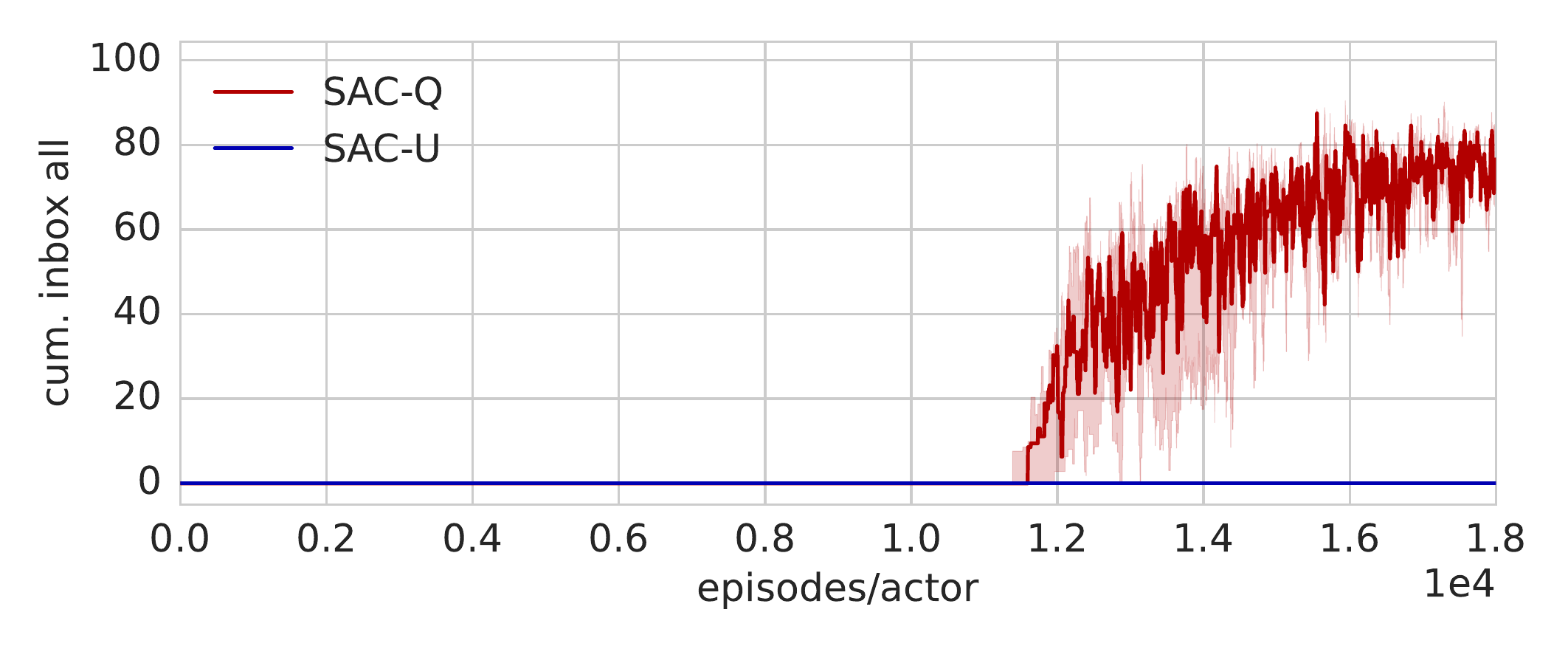}}
\caption{Cleanup experiment, SAC-Q learns all six extrinsic tasks reliably. In addition it reliably learns also to solve the 15 auxiliary tasks in parallel. Part 4: extrinsic tasks.}
\label{cleanup-sac-all-4}
\end{center}
\vskip -0.2in
\end{figure}

A full set of plots for the clean-up tasks is also shown in Figures \ref{cleanup-sac-all-1} to \ref{cleanup-sac-all-4}, comparing the SAC-U and SAC-Q results over all auxiliaries and extrinsic tasks.
While SAC-Q and SAC-U both learn all tasks, only SAC-Q manages to learn the most difficult sparse clean-up task. As shown in the plots, the learned scheduler is more efficient in learning the auxiliaries, as well as the extrinsic tasks, at least in the beginning of the learning process. In later stages, SAC-Q will try to concentrate on intentions that will help it solve the extrinsic tasks, and therefore may disregard some of the less important auxiliaries (e.g. CLOSE(1,2)).


\end{document}